\documentclass[10pt,twocolumn,letterpaper]{article}

\usepackage{iccv}              % To produce the CAMERA-READY version
\usepackage{soul}

\usepackage[utf8]{inputenc}
\usepackage{booktabs}

\definecolor{iccvblue}{rgb}{0.21,0.49,0.74}
\usepackage[pagebackref,breaklinks,colorlinks,allcolors=iccvblue]{hyperref}

\usepackage{multirow}
\usepackage[normalem]{ulem}
\useunder{\uline}{\ul}{}
\usepackage{float}
\usepackage[accsupp]{axessibility}
\def\paperID{*****} % *** Enter the Paper ID here
\def\confName{ICCV}
\def\confYear{2025}

\title{V-RoAst: Visual Road Assessment \protect\\
Can VLM be a Road Safety Assessor Using the iRAP Standard?}

\author{
Natchapon Jongwiriyanurak$^{1}$\thanks{Corresponding author: \href{mailto:natchapon.jongwiriyanurak.20@ucl.ac.uk}{natchapon.jongwiriyanurak.20@ucl.ac.uk}} ,
Zichao Zeng$^{1}$,
June Moh Goo$^{1}$,
Xinglei Wang$^{1}$,
Ilya Ilyankou$^{1}$,\\
Kerkritt Sriroongvikrai$^{2}$,
Nicola Christie$^{1}$,
Meihui Wang$^{1}$,
Huanfa Chen$^{1}$,
James Haworth$^{1}$ \\
$^{1}$ University College London 
$^{2}$ Chulalongkorn University \\
}

\begin{document}
\maketitle
\begin{abstract}
Road safety assessments are critical yet costly, especially in Low- and Middle-Income Countries (LMICs), where most roads remain unrated. 
Traditional methods require expert annotation and training data, while supervised learning-based approaches struggle to generalise across regions. 
In this paper, we introduce \textit{V-RoAst}, a zero-shot Visual Question Answering (VQA) framework using Vision-Language Models (VLMs) to classify road safety attributes defined by the iRAP standard. 
We introduce the first open-source dataset from ThaiRAP, comprising over 2,000 curated street-level images from Thailand, annotated for this task.
We evaluate Gemini-1.5-flash and GPT-4o-mini on this dataset and benchmark their performance against VGGNet and ResNet baselines.
While VLMs underperform on spatial awareness, they generalise well to unseen classes and offer flexible prompt-based reasoning without retraining. Our results show that VLMs can serve as automatic road assessment tools when integrated with complementary data. 
This work is the first to explore VLMs for zero-shot infrastructure risk assessment and opens new directions for automatic, low-cost road safety mapping. 
Code and dataset: \url{https://github.com/PongNJ/V-RoAst}.
\end{abstract}

\section{Introduction}

\begin{figure}[!ht]
    \centering
    \includegraphics[width=.87\linewidth]{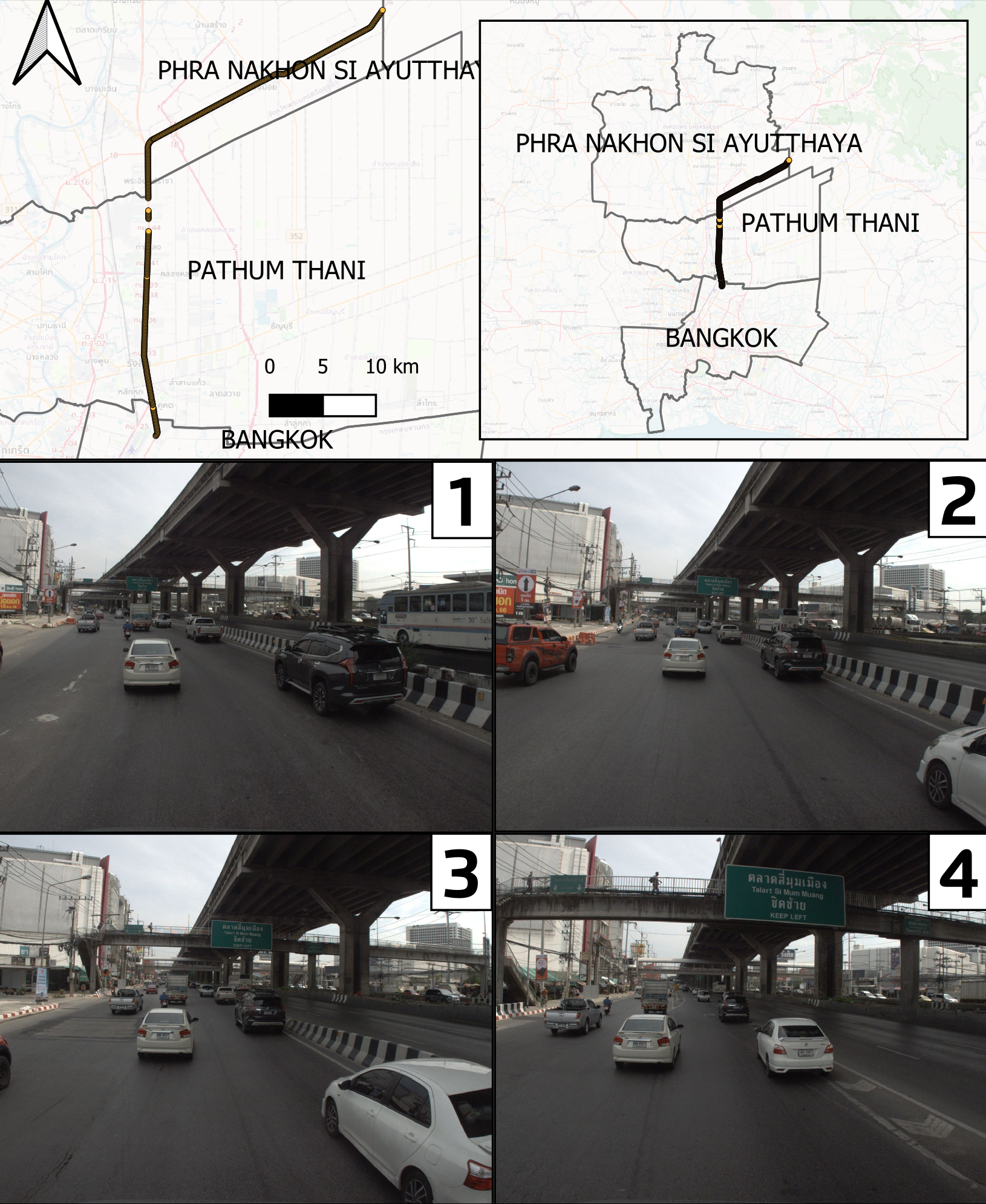}
    \caption{Locations of the ThaiRAP dataset and an example of images in a road segment. The numbers on the images indicate their order within the section.}
    \label{fig:location_dataset}
\end{figure}

Road crashes are a major contributor to global mortality, disproportionately affecting Low- and Middle-Income Countries (LMICs) \citep{who_global_2023, chen_global_2019}. Scalable, low-cost assessment methods are crucial to meeting road safety targets under the UN’s Global Plan for the Decade of Action aiming to ensure that all new roads are built to achieve a rating of at least 3 stars according to the International Road Assessment Programme (iRAP) standard, which rates roads on a scale from 1 to 5, with 5 indicating a safe road and 1 indicating an unsafe road \citep{who_global_2023}. Furthermore, another objective is to improve 75\% of existing roads to more than 3 stars by 2030.

The workflow of iRAP involves road surveys, coding attributes, developing the model, and analysing the results. From iRAP survey manual\footnote{\href{https://resources.irap.org/Specifications/iRAP_Survey_Manual.pdf}{https://resources.irap.org/Specifications/iRAP\_Survey\_Manual.pdf}}, the process requires vehicles and sensors to capture accurate georeferenced images for coding. Once georeferenced images are obtained, trained coders must examine and classify the images according to the codebook manual\footnote{\href{https://resources.irap.org/Specifications/iRAP_Coding_Manual_Drive_on_Right.pdf}{https://resources.irap.org/Specifications/iRAP\_Coding\_Manual\_Drive
\_on\_Right.pdf}}, which requires training and experience. Currently, the road ratings cover mostly highways, as the high cost of surveys has limited broader coverage. Importantly, it is almost impossible for LMICs to assess all roads following this standard. This leaves the vast majority of the road network unrated, making it difficult to reveal the infrastructure risk factors contributing to road deaths. 

To reduce the cost of manual road assessments, automated detection of road features from imagery has become a widely adopted approach, typically using Convolutional Neural Networks (CNNs). While more affordable than human labelling, these models require task-specific training data, making them time-consuming to scale across regions due to visual variability in road environments across cities and countries \citep{kacan_multi-task_2020, pubudu_sanjeewani_learning_2019, jan_convolutional_2018, sanjeewani_single_2021, sanjeewani_optimization_2021}. Other studies have explored alternative data sources to overcome these limitations, including LiDAR \citep{brkic_automatic_2022}, satellite imagery \citep{brkic_utilizing_2023, abdollahi_deep_2020}, UAV imagery \citep{brkic_analytical_2020}, and GPS traces \citep{yin_multimodal_2023}.

Recent studies have explored Visual Language Models (VLMs) for various vision tasks, leveraging their ability to perform without additional training due to large-scale image-text pretraining. Techniques like prompt engineering, Retrieval-Augmented Generation (RAG) \citep{lewis_retrieval-augmented_2021}, and fine-tuning \citep{xing_survey_2024} have been proposed to enhance their performance. VLMs have shown promise in zero-shot tasks such as building age classification \citep{zeng_zero-shot_2024}, landscape image tagging \citep{ilyankou_clip_2025}, and motorcycle risk assessment \citep{jongwiriyanurak_framework_2023}. However, their potential for iRAP-based road attribute classification remains unexplored.

This work investigates VLMs for supporting road assessments by developing prompts to classify iRAP attributes by mimicking a coder observing an image and categorising the attributes, as described in the codebook manual. Additionally, we evaluate the feasibility of leveraging the zero-shot capabilities of VLMs for this task, assuming the models are sufficiently advanced for practical application. In summary, we state the main contributions of our work as follows:
\begin{itemize}
    \setlength\itemsep{0pt} % Adjust the spacing between items
    \setlength\parskip{0pt}  % Adjust the paragraph spacing
    \item \textbf{Open-Source VLM Benchmark}: We propose a new image classification task for VLMs with a real-world open-source dataset from ThaiRAP.
    \item \textbf{Prompt Optimisation}: We optimise the prompts and evaluate the potential of using VLMs to code road attributes using Gemini-1.5-Flash and GPT-4o-mini compared to traditional computer vision models.
    \item \textbf{Zero-Shot Classification}: We demonstrate that VLMs can classify iRAP attributes in a zero-shot setting, achieving competitive performance without any task-specific training. VLMs excel at generalising to unseen images and offer strong potential for further improvement through in-context learning or lightweight fine-tuning.
    \item \textbf{Automatic Road Assessment Framework}: We present an automatic approach using crowdsourced imagery from Mapillary to estimate star ratings.
\end{itemize}

% VLMs achieve competitive results without task-specific training, generalising well to unseen images. Accuracy can be further improved via in-context learning or fine-tuning.
\section{Literature review}
\paragraph{Automated Road Attribute Classification}

Computer vision has long been used to support road safety analysis, particularly in detecting surface defects (e.g., cracks, potholes) and classifying road features \cite{jan_convolutional_2018, ibragimov_automated_2022,goo_hybrid-segmentor_2025,ma_computer_2022}. Several studies have attempted to automate iRAP attribute extraction using CNN-based models trained on labelled street-level imagery \cite{kacan_multi-task_2020, sanjeewani_single_2021,song_farsa_2018}. These approaches, while effective in structured environments, require extensive training data and struggle to generalise across regions due to visual variation and data imbalance \cite{hendrycks_baseline_2018}.

To reduce annotation effort, multi-task learning frameworks have been applied to simultaneously predict several attributes \cite{kacan_dynamic_2024,kacan_multi-task_2020}. However, such models often require domain adaptation techniques to maintain performance across geographic regions \cite{arya_rdd2022_2024, lin_da-rdd_2023, kim_multi-target_2023} and are limited to predefined label sets \cite{kacan_dynamic_2024}. Most can fail to scale to under-resourced contexts, where training data is scarce or unavailable \cite{kacan_dynamic_2024}.

\paragraph{Visual Language Models}

Recent advancements in large Vision-Language Models (VLMs) such as Gemini \cite{team_gemini_gemini_2024}, GPT-4o \cite{openai_gpt-4o_2024}, and LLaVA \cite{liu_visual_2023} have shown promise in zero-shot classification, captioning, and visual reasoning. These models are pretrained on large-scale image-text pairs and can generate responses conditioned on natural language prompts, enabling adaptation to new tasks without fine-tuning.

VLMs have been applied to various spatial tasks, including building understanding and captioning \cite{goo_zero-shot_2024,liang_openfacades_2025}, and autonomous driving \cite{wen_road_2023, jain_semantic_2024,zhang_chatscene_2024,xu_drivegpt4-v2_2025}. While their use in urban analytics is growing, applications in policy-driven, structured classification tasks, such as iRAP-based road assessments, remain largely unexplored.

\paragraph{Visual Question Answering (VQA)}
VQA was initially introduced as a new task in the computer vision and natural language processing domains \citep{agrawal_vqa_2016}. The task involves answering open-ended questions based on images. Several datasets are commonly used to evaluate models, including GQA \citep{hudson_gqa_2019}, OK-VQA \citep{marino_ok-vqa_2019}, A-OKVQA \citep{marino_ok-vqa_2019}, and MMMU \citep{hendrycks_measuring_2021}.  

In addition, VQA datasets focusing on autonomous vehicles, such as KITTI \citep{geiger_vision_2013} and NuScenes-QA \citep{qian_nuscenes-qa_2024}, have been used to advance the field. Jain et al. \cite{jain_semantic_2024} found that GPT-4 performs well in robust driving scenarios that require semantic understanding. Similarly, Dihan et al. \cite{dihan_mapeval_2025} reported that Gemini‑1.5‑Pro demonstrated competitive performance on map-based geospatial reasoning tasks, highlighting its potential in urban analytics applications.

This work utilises a real-world dataset to construct a new image classification benchmark specifically designed for VLMs. Rather than training a new VLM, we explore the potential of existing general-purpose VLMs (e.g., Gemini and GPT) to serve as road assessment assessors in a zero-shot setting. Although non-open-source models incur some usage costs, our goal is to demonstrate that road assessment tasks can benefit from the reasoning capabilities of existing VLMs. This setup also opens possibilities for prompt engineering, RAG, and lightweight fine-tuning to enhance performance.

\section{Dataset Construction}\label{ThaiRAP_Dataset}
\subsection{Data Collection}
We provide a real-world iRAP-compliant road assessment dataset comprising 2,037 street-level images (1600×1200 pixels) captured across Bangkok, Pathum Thani, and Phra Nakhon Si Ayutthaya. The images represent 519 road segments, with 1–4 images per 100\,m segment, and are annotated by trained iRAP coders. Figure~\ref{fig:location_dataset} shows the geographic distribution. Unlike most iRAP datasets, this dataset is publicly available to support research in automated road safety assessment.

\subsection{Dataset Annotation}

\begin{figure}[!t]
    \centering
    \includegraphics[width=1\linewidth]{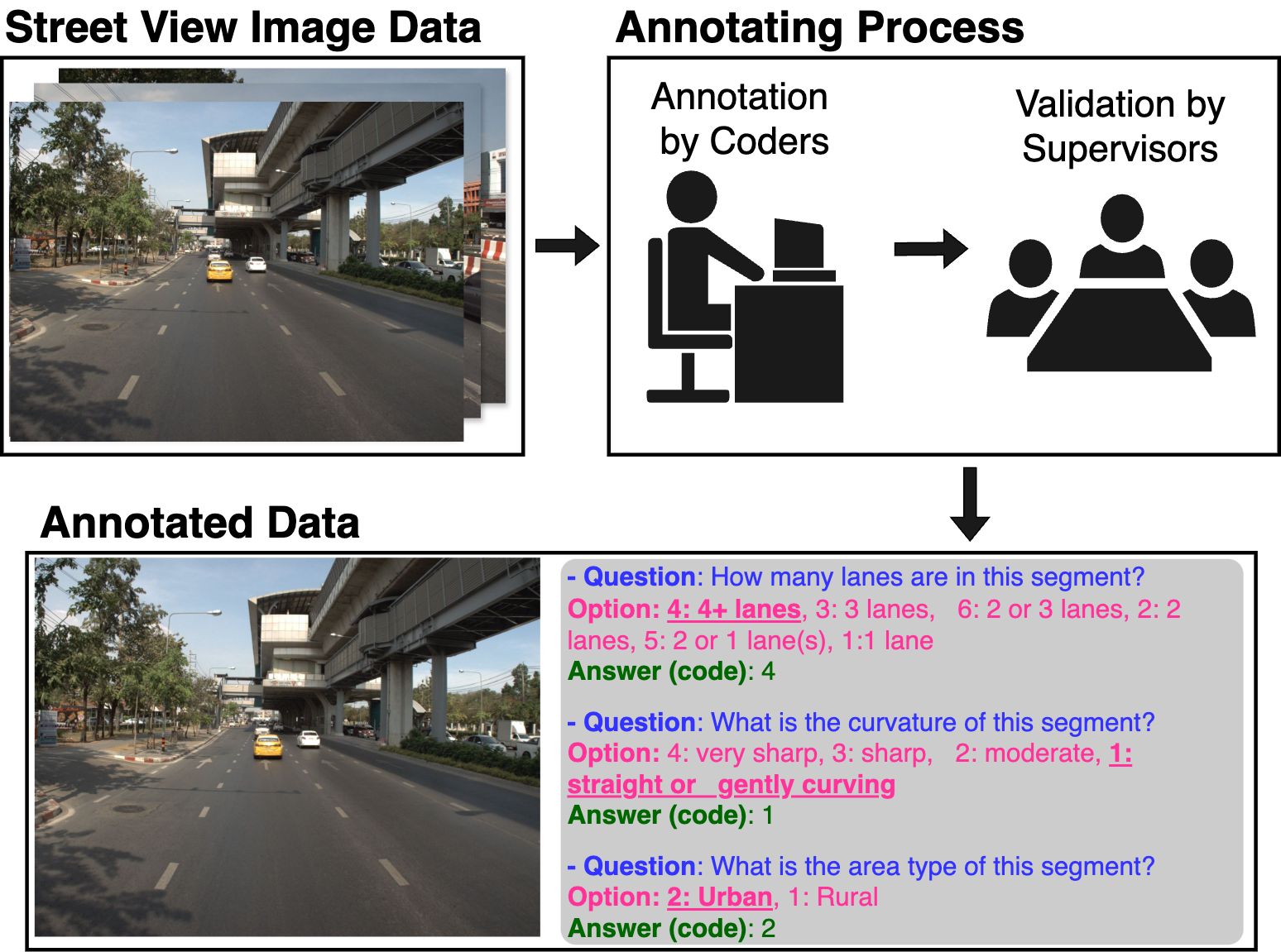}
    \caption{V-RoAst Dataset Annotation Process}
    \label{fig:annotation}
\end{figure}

\begin{figure*}[t]
% \begin{adjustwidth}{-2in}{0in} % shift left by 2.75in
    \centering
    \includegraphics[width=.9\linewidth]{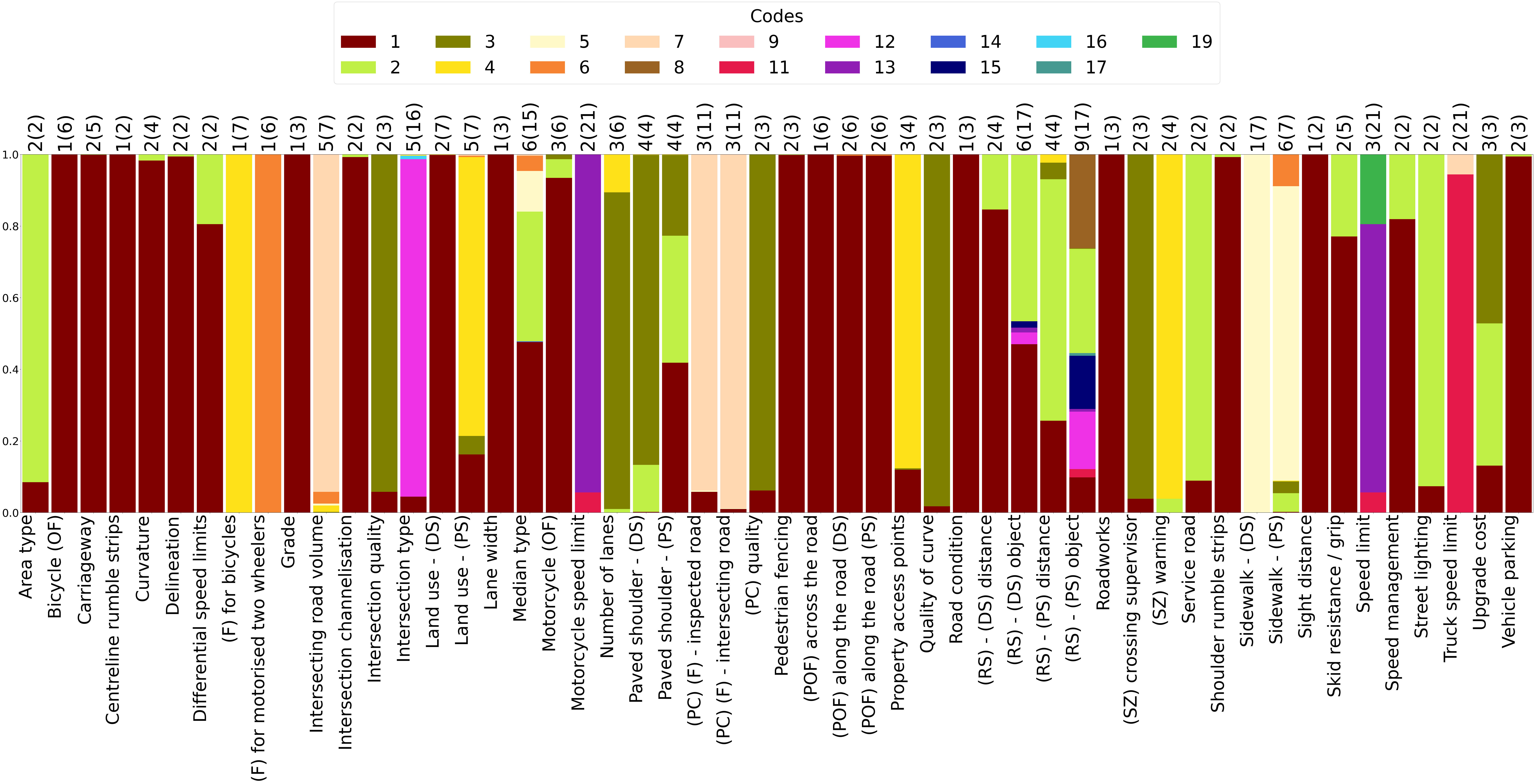}
    \caption{Code distribution: the numbers at the top indicate the unique codes (representing all possible codes). The following abbreviations are used: (OF) = Observed Flow, (F) = Facilities, (DS) = Driver-Side, (PS) = Passenger-Side, (PC) = Pedestrian Crossing, (POF) = Pedestrian Observed Flow, (RS) = Roadside Severity, (SZ) = School Zone.}
    \label{fig:code_distribution}
    % \end{adjustwidth}
\end{figure*}

Each road segment is labelled with 52 iRAP-defined attributes, including number of lanes, road curvature, roadside hazards, and pedestrian facilities. Labels were assigned by trained road safety engineers, following iRAP’s global coding protocol. Figure~\ref{fig:annotation} illustrates the annotation process, and Figure~\ref{fig:code_distribution} shows class distribution across all coded attributes. The dataset includes both categorical and ordinal attributes, enabling fine-grained evaluation of classification performance across imbalanced classes.

\subsection{Dataset Statistics}

The dataset includes ordinal attributes, enabling fine-grained evaluation of classification models. However, the number of classes per attribute varies (shown at the top of each bar in Figure~\ref{fig:code_distribution}). The dataset exhibits strong class imbalance: 11 attributes contain only a single class, making them unusable for supervised training, while several others include rare classes with fewer than 10 samples.

Not all attributes span the full set of possible iRAP codes(classes). For example, ``Bicycle Observed Flow'' contains only 6 of the official classes in this dataset, and ``Centreline Rumble Strips'' appears with just 2. These limitations highlight the practical constraints of real-world data collection and motivate the use of zero-shot classification approaches that do not require balanced training data.

\subsection{Data Preprocessing}

To compare our method with traditional computer vision baselines, the dataset (n = 2,037) was divided into training (1,274 original + 464 augmented), testing (492), validation (243), and unseen (28) sets. We ensured class-balanced splits for each attribute where possible. The splitting followed these rules:

\begin{enumerate}
    \item Attributes with only one class were excluded.
    \item Classes with 5--11 samples were augmented.
    \item Classes with 4 or fewer samples were augmented only if the attribute had two classes.
    \item If an attribute had more than two classes and a class contained $\leq$ 4 samples, those samples were moved to the unseen set.
\end{enumerate}

To mitigate class imbalance (Rules 2 and 3), five types of noise were applied to selected images: Gaussian, salt-and-pepper, speckle, periodic, and quantisation noise. All augmented images were included only in the training set.

The unseen set was reserved for evaluating zero-shot prediction performance. Baseline models were trained using the training set and validated on the validation set. Final evaluations for both baselines and VLMs were conducted on the testing and unseen sets.

\section{Proposed Method}
\subsection{Problem Definition}
Given an input consisting image $I$ and associated metadata $M$ including image ID, latitude, and longitude, the objective is to accurately classify it across 52 distinct attribute types, where each attribute type is treated as a separate multi-class classification problem. The \textbf{V}isual \textbf{Ro}ad \textbf{As}sessmen\textbf{t} (V-RoAst) framework incorporates a vision-language model $f$, guided by a system prompt $T_S$ and a user prompt $T_U$, to perform the classification. The output is a set of 52 predictions $\mathcal{A} = \{a_1, a_2,...a_{52}\}$, where $\mathcal{A}$ denotes the set of predicted attributes,  $a_i \in C_i$, and $C_i$ is the set of possible classes for the $i$-th attribute type.

% The evaluation is conducted at the road level, rather than at the individual image level. 
For each road segment, we are provided with between 1 and 4 images, each associated with its own set of predicted attributes. The goal is to derive a single aggregated attribute set per road that reflects the most critical conditions, consistent with road safety assessment standards.
Specifically, for each attribute type, each image produces a prediction $a_i^{j} \in \mathcal{A}$, where $j \in \{1,...,n\} (n\leq4)$  indexes the images for a given road. The predicted attribute $\hat{a}_i^{r_k}$ for the road $r_k$ is selected as the highest-risk class among the image-level predictions, according to a predefined risk ranking from the iRAP specification. The aggregated road-level preditions $\hat{\mathcal{A}}^{r_k}=\{\hat{a}_1^{r_k},...,\hat{a}_{52}^{r_k}\}$ and then compared against the ground truth road attributes $\hat{\mathcal{A}}_g^{r_k}=\{\hat{a}_{g1}^{r_k},...,\hat{a}_{g52}^{r_k}\}$  using standard multi-class classification metrics.

% We adopt a multi-view or feature-fusion strategy to ensure that critical relationships across the entire road segment are captured in a single forward pass. Specifically, for each road segment \(S\) represented by four images \(I_1, I_2, I_3, I_4\) and for each attribute \(j\) (with \(j = 1,\dots,52\)), let \(f_{j,c}(I_i)\) denote the model's score for class \(c \in \mathcal{C}_j\) from image \(I_i\). The final prediction for attribute \(j\) is given by:

% \begin{equation}
% \hat{y}_j = \max_{c \in \mathcal{C}_j} \left\{ \max_{i \in \{1,2,3,4\}} f_{j,c}(I_i) \right\}
% \end{equation}

% This formulation aggregates the scores across the four images by taking the maximum score for each class and then selecting the class with the highest aggregated score. This approach avoids inconsistencies if the images are classified separately and combined, producing more accurate and consistent predictions aligned with iRAP assessment guidelines.

\subsection{\texorpdfstring{\textbf{\underline{V}}isual \textbf{\underline{Ro}}ad \textbf{\underline{As}}sessmen\textbf{\underline{t}} Framework}{Visual Road Assessment and Prompt Design}} \label{section_experiment}

\begin{figure}[!t]
    \centering
    \includegraphics[width=1\linewidth]{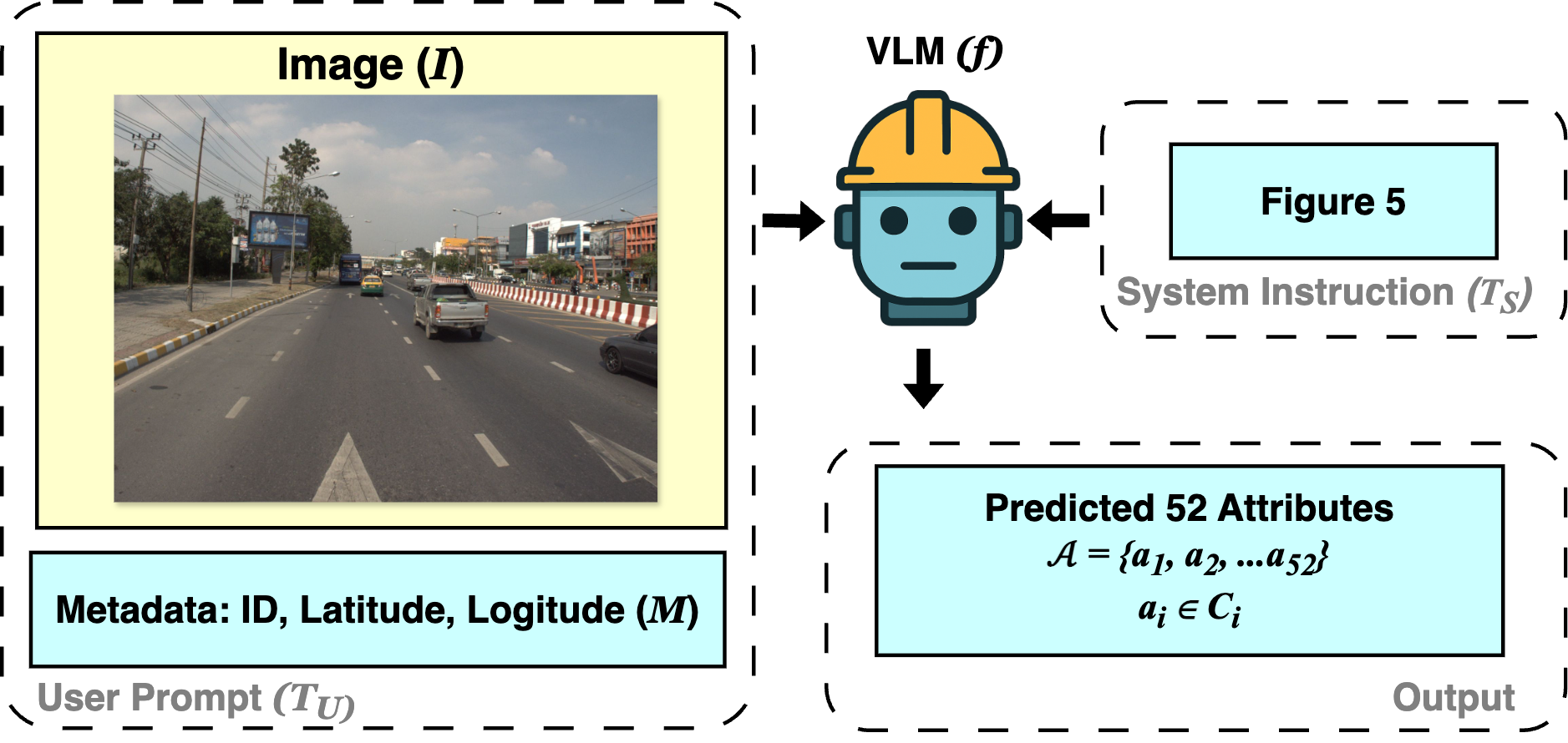}
    \caption{Framework of V-RoAst for Visual Road Assessment}
    \label{fig:V-RoAst-Framework}
\end{figure}

\begin{figure*}[!t]
    \centering
    \includegraphics[width=.85\linewidth]{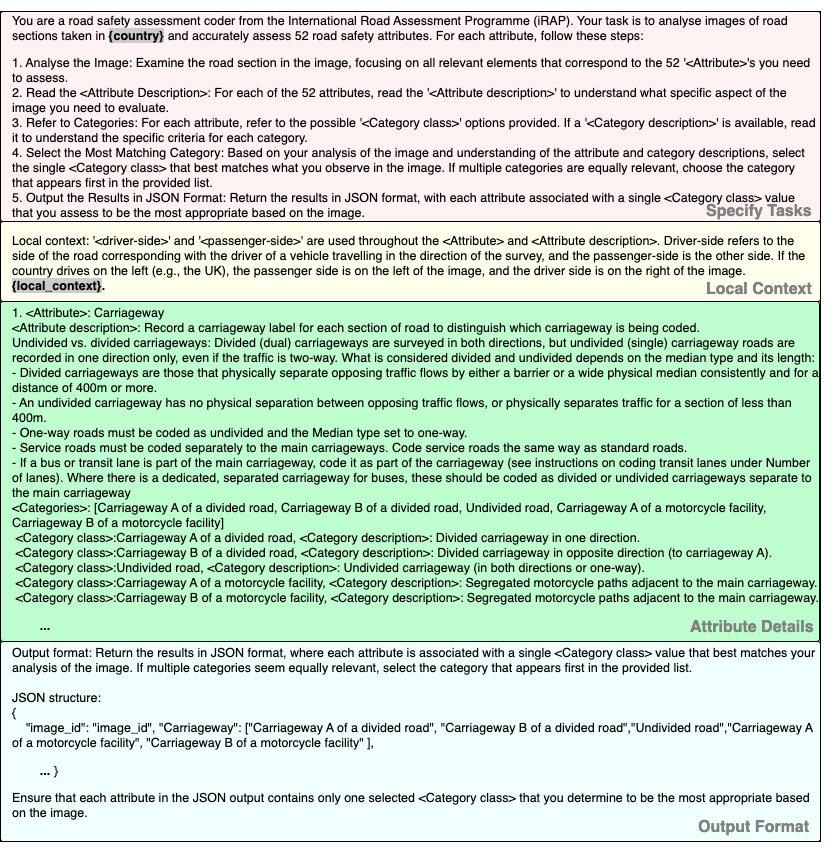}
    \caption{System Prompt from Figure \ref{fig:V-RoAst-Framework}}
    \label{fig:text prompt}
    % \end{adjustwidth}
\end{figure*}

V-RoAst is a framework designed to classify 52 iRAP attributes $\mathcal{A}$ from the input image $I$, as shown in Figure~\ref{fig:V-RoAst-Framework}. This framework is easily applicable in any city, requiring minimal expertise. The framework uses VLMs via text input for system instructions $T_S$ and user prompts $T_U$, which include $I$ and $M$. The workflow operates by inputting an image $I$ and its associated metadata $M$, including task specifications, local context, attribute details, and output format into the VLM $f$. This process enables the generation of attribute-level predictions $\mathcal{A}$ for each image $I$. 
% Figure~\ref{fig:text prompt} shows an example prompt used in the system. 

\subsubsection{System Instruction} %system prompt
As shown in Figure \ref{fig:text prompt}, System Instruction ($T_S$) is divided into 4 parts, including task specification, local context, attribute details, and output format as suggested by OpenAI \cite{openai_gpt-4_2024} and Gemini \cite{team_gemini_gemini_2024} technical report. 

\begin{enumerate}
    \item \textbf{Task Specification:}  
    This component provides step-by-step instructions for processing each image. The model is guided to classify each \texttt{$<$Attribute$>$} by selecting the most appropriate \texttt{$<$Category class$>$} based on its visual description. This section also specifies how the output should be formatted. Notably, \texttt{\{country\}} is included to localise the analysis for each region.

    \item \textbf{Local Context:} Information such as which side of the road vehicles drive on is provided through \texttt{\{local\_context\}}. This helps the VLMs interpret images more accurately by incorporating local traffic rules and infrastructure norms. It is especially useful for visually ambiguous attributes, like the presence of service roads, where local knowledge improves classification. Including this input mimics how human coders use familiarity with the area to make more accurate assessments.

    % Contextual details, such as the side of the road on which vehicles drive (left or right), are provided through \texttt{\{local\_context\}}. This information helps the VLMs interpret images more accurately. For example, attributes like speed limits or specific infrastructure features are clarified using local regulatory details, enhancing the model's relevance and precision.

    % Incorporating \texttt{\{local\_context\}} is particularly important for attributes that are difficult to classify visually, such as the presence of service roads. This additional input of local context enables the system to account for subtle, location-specific nuances, ensuring robustness in its assessments, as this is to mimic how coders know the area.

    \item \textbf{Attribute Details:}  
    Each attribute is defined with its name, description (\texttt{$<$Attribute description$>$}), and possible classes (\texttt{$<$Category class$>$}), accompanied by descriptions of each category (\texttt{$<$Category description$>$}). These structured definitions ensure that the models have clear guidelines for classification. Note that this section includes approximately 100,000 characters (or 20,000 tokens) from the iRAP coding manual, which this work made available in JSON format.

    \item \textbf{Output Format:}  
    The results are returned in JSON format to maintain consistency. The models are instructed to select the best match from the provided \texttt{$<$Category class$>$} list, making the outputs standardised and interpretable.
\end{enumerate}

\subsubsection{User Prompt}
In this experiment, the User Prompt ($T_U$) combines both visual input and accompanying metadata, including \texttt{\{image\_id\}} and \texttt{\{latitude, longitude\}}. These metadata elements help the vision–language models (VLMs) format their outputs more consistently and provide geographic context to guide interpretation.

Speed-related attributes such as “Speed Limit” and “Motorcycle Speed Limit” are essential for iRAP assessments but are often not directly observable from imagery alone, particularly when speed signage is absent or unclear. In V-RoAst, these attributes are predicted by providing the VLM with both the image and its coordinates in the prompt. The location information enables the model to draw on spatial priors learned during pretraining (e.g., typical speed regulations in that region) while also inferring additional contextual cues, such as road type, land use, and traffic environment, from the image itself. By combining these two sources, the model can estimate the most likely speed limit even when explicit signage is missing. For example, two visually similar rural roads may have different statutory limits depending on the province; the geographic cue guides the model toward the correct classification. This approach improves the contextual relevance and accuracy of speed-related predictions without requiring explicit annotation of road type or land use.

% For example, integrating local speed limit laws allows the VLMs to accurately assess speed-related attributes, such as motorcycle speed limits, truck speed limits, and differential speed limits. This integration ensures that the models consider both visual cues and regulatory context, improving the relevance of the analysis to the local environment and regulation.

% Since VLMs are pretrained on massive, diverse datasets that often include geospatial and regulatory patterns, incorporating location-aware metadata allows them to draw on this broader knowledge. This multimodal design—combining visual scenes with contextual text—enables the model to reason more effectively, particularly for attributes that may not be visually obvious. Overall, this dual-input prompting improves both the accuracy and contextual relevance of the model’s assessments.

% By leveraging both visual and textual data, the VLMs can better interpret complex scenes, even in cases where certain attributes are less visually apparent. This dual-input approach enhances the accuracy and reliability of the assessments.

\subsection{Vision Language Models}

This work evaluates Gemini-1.5-flash and GPT-4o-mini as $f$ for their potential to replicate the work of road safety assessors under the iRAP standard. These models require no additional training or significant computational resources, making them accessible for use by local stakeholders. As demonstrated by Yue et al. \cite{yue_mmmu_2024}, Gemini and GPT outperform other models in the Multi-discipline Multimodal Understanding and Reasoning benchmark. The experiments were conducted through the Gemini and the OpenAI APIs.

To assess their performance, we analysed 337 road segments (1348 images) with Gemini-1.5-flash and GPT-4o-mini, alongside ResNet and VGGNet baseline models. An additional set of 7 segments with unseen attributes was used to evaluate zero-shot classification capabilities. 

\subsection{Image Processor for Mapillary Imagery}
Crowdsourced Street View Images (SVIs) are accessible on various platforms, with Mapillary being one of the most well-known, providing an API for image downloads. Hence, we evaluate the system using the V-RoAst framework with Gemini-1.5-flash to classify all 52 attributes. Then, we combine these results with additional information, including operating speed and ADDT from the ground truth, to determine the star rating outcomes.

For this work, images were obtained using a 50-metre buffer around ThaiRAP locations under the condition that the images were captured within one year of the collecting date. However, only 42 road segments were found to have corresponding Mapillary images, yielding 165 images. Panoramic images were converted to 1200x1600 binocular view images to align with the ThaiRAP data format. %Subsequently, these images were processed using V-RoASt to examine attributes, as explained in \ref{section_experiment}. 

It is important to note that some attributes may differ from the iRAP ground truth. For example, the number of vehicles shown in image (observed flow) and the number of vehicles parked in the captured scene can vary. To validate the automatic use of V-RoAst, we did not verify whether the downloaded images originated from the same road, which could potentially affect the predicted star rating.
\section{Experiments}
\subsection{Implementation Detail}
\paragraph{Baseline} %after this use classes = codes, classify = code (v./ task). 
Our work used VGGNet and ResNet as baseline models to compare with our proposed approach. 
These models were selected because they have been widely used in previous work on road safety assessment and multi-task classification \citep{song_farsa_2018, kacan_multi-task_2020,kacan_dynamic_2024}.
Since they are designed for single-task classification, we adapt the architectures for a multi-attribute coding problem, where a single encoder is shared across all tasks, and separate decoders are allocated for each individual task. 
\begin{itemize}
    \item \textbf{VGGNet} \citep{simonyan_very_2015} is a deep neural network model known for its simple architecture that stacks multiple convolutional layers with small 3x3 filters, achieving high performance in image classification tasks and becoming a widely used baseline in computer vision.
    \item \textbf{ResNet} \citep{he_deep_2015} is also a deep convolutional neural network that utilises residual blocks and skip connections to enhance feature learning at various abstraction levels, making it highly effective for image classification and transfer learning tasks.
\end{itemize}

\paragraph{Evaluation Metrics}
We report results at two granularities. \textbf{Overall} performance is the macro-average of each metric across the iRAP attribute groups (“sections”), summarising model behaviour over broad feature types.  \textbf{Attribute-level} performance is computed for each of the 52 individual attributes, revealing fine-grained strengths and weaknesses. When a road segment carries multiple risk classes, we follow the iRAP convention and keep the first (highest-risk) label. For every attribute, we calculate accuracy, precision, recall, and F1-score, then macro-average them so that attributes with few samples contribute equally to the final score.

\subsection{Results and Discussions}
% We evaluate the performance of V-RoAst using two state-of-the-art Vision-Language Models (VLMs)—Gemini-1.5-flash and GPT-4o-mini—in comparison with supervised baselines (VGGNet and ResNet). 
% The evaluation covers both macro-averaged metrics across all attributes and subgroup analyses.
% Attribute-level results are included in the Appendix.

\subsubsection{Overall Performance}

\begin{table}[!t]
\centering
\caption{
% Macro-averaged Accuracy, Precision, Recall, and F1-score for overall and unseen class performance. 
% \textbf{Bold} highlights the best-performing visual language model (VLM) among VLMs (GPT and Gemini); 
% \underline{Underlined} indicates the best-performing model overall (including vision-only CNNs). 
% \textbf{\underline{Bold and underlined}} denotes the best-performing VLM that also achieves the highest score overall.
Macro-averaged Accuracy, Precision, Recall, and F1-score for overall and unseen class performance. 
\textbf{Bold} highlights the best-performing model.
}
\label{tab:macro_metrics}
\begin{tabular}{llcccc}
\toprule
\textbf{Group Attribute} & \textbf{Model} & \textbf{Acc} & \textbf{Pre} & \textbf{Rec} & \textbf{F1} \\
\midrule
\multirow{4}{*}{All Attributes} 
  & \multicolumn{5}{l}{\textit{CNN Models}} \\
  & \quad VGG         & \textbf{0.96} & 0.76 & 0.74 & 0.75 \\
  & \quad ResNet      & \textbf{0.96} & \textbf{0.88} & \textbf{0.86} & \textbf{0.86} \\
  & \multicolumn{5}{l}{\textit{VLM Models}} \\
  & \quad GPT         & 0.75 & 0.46 & 0.47 & 0.42 \\
  & \quad Gemini      & {0.82} & {0.49} & {0.50} & {0.47} \\
\midrule
\multirow{4}{*}{Unseen Classes} 
  & \multicolumn{5}{l}{\textit{CNN Models}} \\
  & \quad VGG         & 0.24 & 0.18 & 0.23 & 0.18 \\
  & \quad ResNet      & 0.29 & 0.39 & 0.25 & 0.21 \\
  & \multicolumn{5}{l}{\textit{VLM Models}} \\
  & \quad GPT         & 0.48 & 0.34 & 0.36 & 0.34 \\
  & \quad Gemini      & \textbf{{0.62}} & \textbf{{0.45}} & \textbf{{0.46}} & \textbf{{0.43}} \\
\bottomrule
\end{tabular}
\end{table}

% \begin{table}[!t]
% \centering
% \caption{Macro-averaged Accuracy, Precision, Recall, and F1-score for overall performance and unseen class generalisation. \textbf{Bold} denotes the best-performing visual language model (VLM); \underline{underlined} highlights the best-performing model.}
% \label{tab:macro_metrics}
% \begin{tabular}{llcccc}
% \toprule
% \textbf{Group Attribute} & \textbf{Model} & \textbf{Acc} & \textbf{Pre} & \textbf{Rec} & \textbf{F1} \\
% \midrule
% % \multicolumn{6}{l}{\textit{Overall Performance}} \\
% % \midrule
% \multirow{4}{*}{All Attributes} 
%   & VGG         & \underline{0.96} & 0.76 & 0.74 & 0.75 \\
%   & ResNet      & \underline{0.96} & \underline{0.88} & \underline{0.86} & \underline{0.86} \\
%   & GPT         & 0.75 & 0.46 & 0.47 & 0.42 \\
%   & Gemini      & \textbf{0.82} & \textbf{0.49} & \textbf{0.50} & \textbf{0.47} \\
% \midrule
% \multirow{4}{*}{Unseen Classes} 
%   & VGG         & 0.24 & 0.18 & 0.23 & 0.18 \\
%   & ResNet      & 0.29 & 0.39 & 0.25 & 0.21 \\
%   & GPT         & 0.48 & 0.34 & 0.36 & 0.34 \\
%   & Gemini      & \textbf{\underline{0.62}} & \textbf{\underline{0.45}} & \textbf{\underline{0.46}} & \textbf{\underline{0.43}} \\
% \bottomrule
% \end{tabular}
% \end{table}

Table~\ref{tab:macro_metrics} presents the macro-averaged accuracy, precision, recall, and F1-score across all 52 attributes. Among the supervised baselines, ResNet achieves the highest macro performance with an accuracy of 0.96, precision of 0.88, recall of 0.86, and F1-score of 0.86. VGGNet also performs strongly, with slightly lower recall and precision, indicating solid but less consistent classification across attributes.

In contrast, the zero-shot VLMs show reduced performance. Gemini-1.5-flash achieves a macro-accuracy of 0.82 and an F1-score of 0.47, while GPT-4o-mini performs slightly behind. Despite this performance gap, VLMs offer a crucial advantage, the ability to predict all possible iRAP attribute classes without requiring task-specific training data. This is particularly valuable in data-sparse or label-scarce settings.

On the other hand, the \textit{Unseen Classes},  those with highly imbalanced or missing training data (and thus excluded from the supervised baselines), VLMs excel. Gemini-1.5-flash achieves a macro-accuracy of 0.62 and an F1-score of 0.43, demonstrating its potential for generalised attribute classification without training.

\begin{table}[!t]
\centering
\caption{
Macro-averaged Accuracy, Precision, Recall, and F1-score across iRAP-defined group attributes. 
\textbf{Bold} highlights the best-performing model.
}
\label{tab:group_metrics}
\begin{tabular}{llcccc}
\toprule
\textbf{Group Attribute} & \textbf{Model} & \textbf{Acc} & \textbf{Pre} & \textbf{Rec} & \textbf{F1} \\
\midrule
\multirow{6}{*}{Observed Flows} 
  & \multicolumn{5}{l}{\textit{CNN Models}} \\
  & \quad VGG         & \textbf{0.99} & 0.83 & 0.83 & 0.83 \\
  & \quad ResNet      & \textbf{0.99} & \textbf{0.89} & \textbf{0.99} & \textbf{0.91} \\
  & \multicolumn{5}{l}{\textit{VLM Models}} \\
  & \quad GPT         & 0.95 & 0.47 & 0.49 & 0.47 \\
  & \quad Gemini      & 0.98 & 0.68 & 0.65 & 0.66 \\
\midrule
\multirow{6}{*}{Speed Limits} 
  & \multicolumn{5}{l}{\textit{CNN Models}} \\
  & \quad VGG         & \textbf{0.99} & 0.72 & 0.73 & 0.73 \\
  & \quad ResNet      & 0.98 & \textbf{0.98} & \textbf{0.98} & \textbf{0.98} \\
  & \multicolumn{5}{l}{\textit{VLM Models}} \\
  & \quad GPT         & 0.52 & 0.46 & 0.52 & 0.34 \\
  & \quad Gemini      & 0.84 & 0.55 & 0.53 & 0.53 \\
\midrule
\multirow{6}{*}{Mid-block} 
  & \multicolumn{5}{l}{\textit{CNN Models}} \\
  & \quad VGG         & 0.97 & 0.67 & 0.64 & 0.64 \\
  & \quad ResNet      & \textbf{0.98} & \textbf{0.92} & \textbf{0.88} & \textbf{0.90} \\
  & \multicolumn{5}{l}{\textit{VLM Models}} \\
  & \quad GPT         & 0.77 & 0.59 & 0.58 & 0.52 \\
  & \quad Gemini      & 0.85 & 0.60 & 0.64 & 0.59 \\
\midrule
\multirow{6}{*}{Roadside} 
  & \multicolumn{5}{l}{\textit{CNN Models}} \\
  & \quad VGG         & 0.90 & 0.72 & \textbf{0.73} & 0.72 \\
  & \quad ResNet      & \textbf{0.91} & \textbf{0.76} & 0.72 & \textbf{0.74} \\
  & \multicolumn{5}{l}{\textit{VLM Models}} \\
  & \quad GPT         & 0.33 & 0.27 & 0.22 & 0.18 \\
  & \quad Gemini      & 0.36 & 0.20 & 0.21 & 0.17 \\
\midrule
\multirow{6}{*}{Intersections} 
  & \multicolumn{5}{l}{\textit{CNN Models}} \\
  & \quad VGG         & 0.95 & 0.76 & 0.73 & 0.74 \\
  & \quad ResNet      & \textbf{0.97} & \textbf{0.82} & \textbf{0.77} & \textbf{0.79} \\
  & \multicolumn{5}{l}{\textit{VLM Models}} \\
  & \quad GPT         & 0.88 & 0.46 & 0.48 & 0.45 \\
  & \quad Gemini      & 0.90 & 0.49 & 0.48 & 0.46 \\
\bottomrule
\end{tabular}
\end{table}

% \begin{table}[!t]
% \centering
% \caption{Macro-averaged accuracy, precision, recall, and F1-score for overall performance and unseen class generalisation.
% \textbf{Bold} highlights the best-performing vision-language model (VLM); \underline{underlined} values indicate the overall best performance across all models.}
% \label{tab:macro_metrics}
% \begin{tabular}{llcccc}
% \toprule
% \textbf{Group Attribute} & \textbf{Model} & \textbf{Acc} & \textbf{Pre} & \textbf{Rec} & \textbf{F1} \\
% \midrule
% % \multicolumn{6}{l}{\textit{Overall Performance}} \\
% % \midrule
% \multirow{4}{*}{All Attributes} 
%   & VGG         & \underline{0.96} & 0.76 & 0.74 & 0.75 \\
%   & ResNet      & \underline{0.96} & \underline{0.88} & \underline{0.86} & \underline{0.86} \\
%   & GPT         & 0.75 & 0.46 & 0.47 & 0.42 \\
%   & Gemini      & \textbf{0.82} & \textbf{0.49} & \textbf{0.50} & \textbf{0.47} \\
% \midrule
% \multirow{4}{*}{Unseen Classes} 
%   & VGG         & 0.24 & 0.18 & 0.23 & 0.18 \\
%   & ResNet      & 0.29 & 0.39 & 0.25 & 0.21 \\
%   & GPT         & 0.48 & 0.34 & 0.36 & 0.34 \\
%   & Gemini      & \textbf{\underline{0.62}} & \textbf{\underline{0.45}} & \textbf{\underline{0.46}} & \textbf{\underline{0.43}} \\
% \bottomrule
% \end{tabular}
% \end{table}

\subsubsection{Performance by Attribute Group}

Subgroup analyses highlight nuanced differences in Table \ref{tab:group_metrics}. For spatially grounded attributes, such as roadside attributes, supervised models outperform VLMs significantly. Their training allows precise estimation of spatial and geometric features.

However, for attributes that are less spatially dependent, such as observed flows or mid-block attributes, VLMs show more competitive performance. Notably, Gemini-1.5-flash consistently outperforms GPT-4o-mini across all attribute groups, suggesting stronger generalisation and reasoning capabilities in visual question answering tasks.

\subsubsection{Attribute-Level Analysis} \label{attribute-level_analysis}

Attribute-level performance analysis shows that baseline models were trained only on attributes with sufficient class diversity, excluding 11 attributes that contained a single class. VLMs are not constrained by this limitation and can generate predictions for all attributes, offering broader applicability. Full attribute-level results are provided in the Supplementary Material.

We observe that VLMs perform better when recognising prominent visual cues (e.g., vehicle parking, area type) but underperform in estimating distances or interpreting scene geometry. These results suggest that while VLMs are effective for attribute presence/absence tasks, they may struggle with precise spatial reasoning \cite{roberts_gpt4geo_2023}.

% Table~\ref{tab:attr_performance_full} in the Appendix illustrates per-attribute performance. Baseline models were trained only on attributes with sufficient class diversity, excluding 11 attributes with a single class. VLMs are not constrained by this limitation and generate predictions for all attributes, offering broader applicability.

% We observe that VLMs perform better when recognising prominent visual cues (e.g., presence of a school zone, number of lanes) but underperform in estimating distances or interpreting scene geometry. These results suggest that while VLMs are effective for attribute presence/absence tasks, they may struggle with precise spatial reasoning \cite{roberts_gpt4geo_2023}.

% \subsubsection{Interpretation and Implications}
% While supervised models offer higher accuracy and consistency, they are limited by their dependence on labelled training data and cannot generalise to unseen attribute classes. VLMs, although less accurate overall, offer flexibility and scalability, making them suitable for rapid deployment in new geographies or applications where annotated data is scarce.

% This trade-off suggests an opportunity for a hybrid approach: leveraging VLMs for scalable zero-shot predictions and pre-labelling, and then using supervised models to fine-tune or validate results for critical or spatially complex attributes. Future work could explore few-shot fine-tuning or in-context learning to further bridge the performance gap between VLMs and supervised methods.

\subsubsection{Qualitative Assessment using VQA}

A key advantage of VLMs lies in their ability to perform VQA, enabling users, regardless of technical background, to iteratively refine model outputs through prompt adaptation. This makes VLMs particularly suitable for participatory or practitioner-led applications, where model interpretability and adaptability are crucial.

\begin{figure}[!t]
    \centering
    \includegraphics[width=1\linewidth]{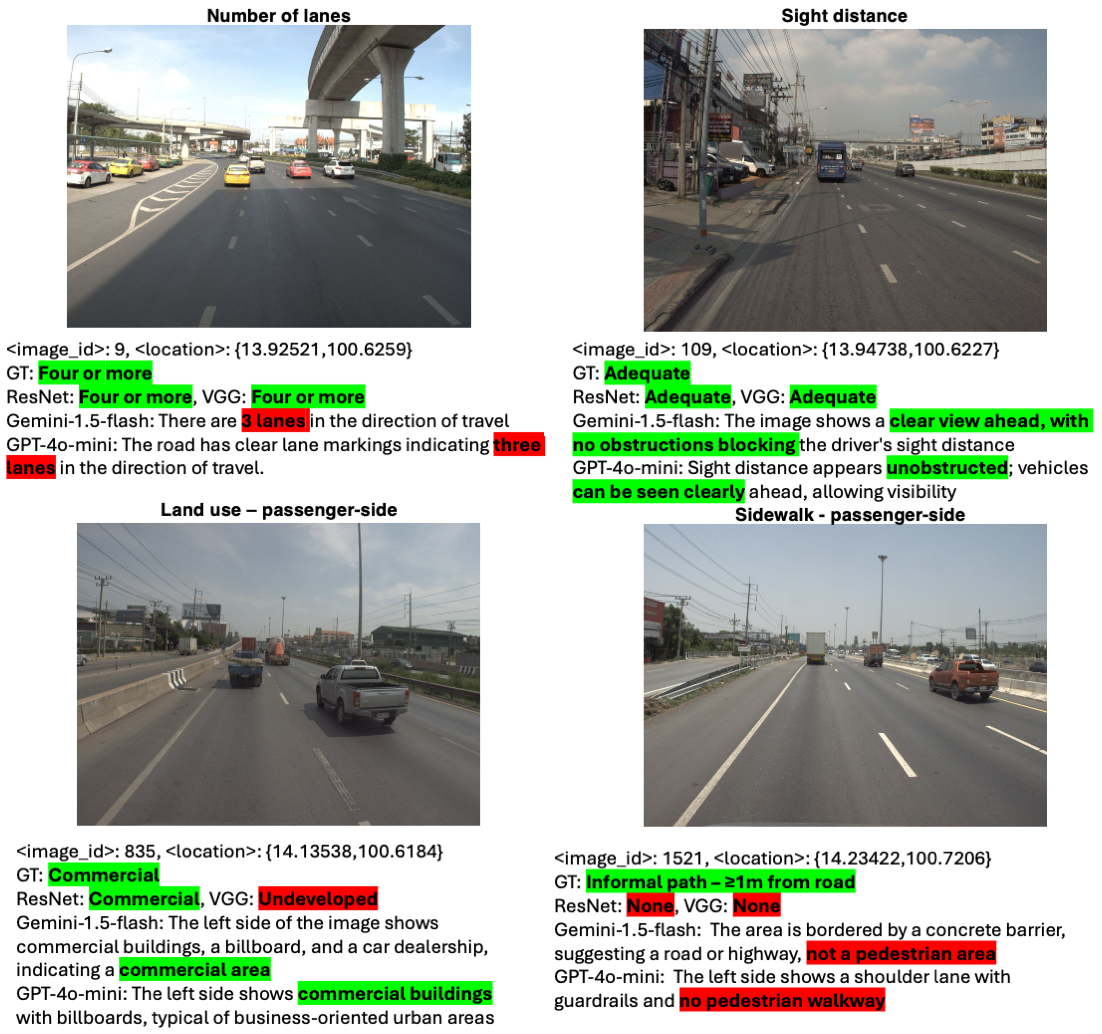}
    \caption{Qualitative assessment of VLM performance using VQA with correct (green) and wrong (red) answers}
    \label{fig:qualitative}
\end{figure}

Figure~\ref{fig:qualitative} illustrates how prompt tuning can be employed to guide model reasoning and address misclassification. For example, in the bottom-right case, none of the four models tested correctly identified the ``Sidewalk - passenger-side'' attribute. The ground truth label referred to an informal path located more than 1 metre from the main carriageway, a feature not clearly visible in the street-view image. This discrepancy highlights a common challenge that visual cues may be insufficient or ambiguous, requiring local contextual knowledge for accurate classification. 

In such cases, supplementing street-level imagery with additional data sources, such as satellite views or local GIS layers, could help disambiguate difficult scenes. The interactive nature of VQA-based models also allows for on-the-fly prompt modifications to explore alternative interpretations, further enhancing usability for road safety assessments in resource-constrained environments.

\subsubsection{Automatic Road Assessment}

\begin{figure}[!t]
    \centering
    \includegraphics[width=.9\linewidth]{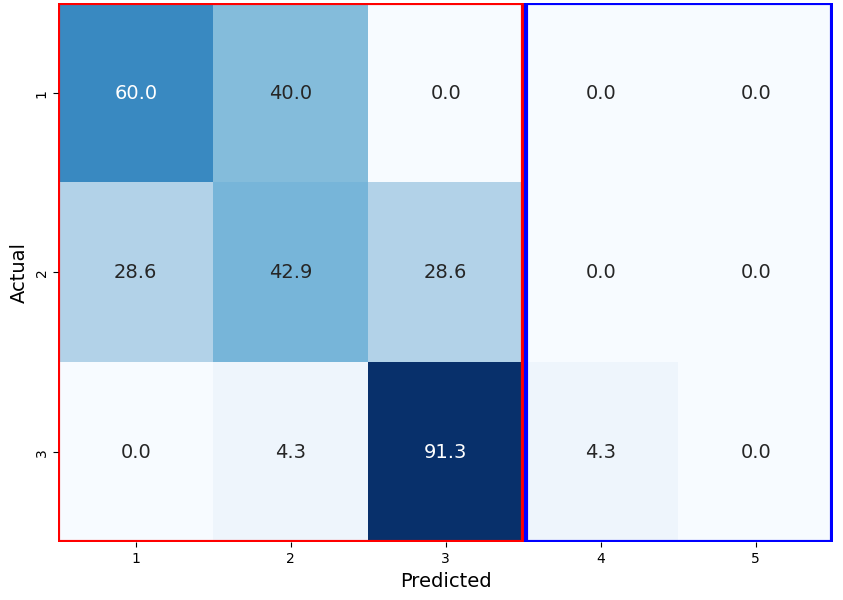}
    \caption{Star rating (motorcyclists) confusion matrix of using crowdsourced imagery with V-RoAst and ground truth from ThaiRAP}
\label{fig:confusion_matrix_star_rating}
\end{figure}

Figure \ref{fig:confusion_matrix_star_rating} presents the confusion matrix for star rating predictions (motorcyclists) using Mapillary images with V-RoAst (Gemini-1.5-flash). The results demonstrate the model's effectiveness in identifying high-risk roads with star ratings below 3, highlighted in the red box.

The flexibility of the V-RoAst framework lies in its integration of \texttt{\{local\_context\}}, allowing local stakeholders to tailor prompts based on their expertise and validate results against ground truth data. This adaptability enables precise identification of high-risk areas, supporting road safety initiatives and investment prioritisation. Although Mapillary's current coverage is limited, its crowdsourced platform serves as a valuable resource that transport authorities can leverage \citep{wang_investigating_2024,hou_global_2024}. Similar applications using commercial platforms like Google Street View may be explored, subject to licensing agreements.

Additional inputs, such as Annual Average Daily Traffic (AADT) and operating speeds, are essential to achieving comprehensive safety assessments. While this information is not derived directly from image labels, they are critical in determining road safety performance and star ratings.

\section{Conclusion and Future work} % Can VLM be a Road Safety Assessor using the iRAP standard?

This work introduced V-RoAst, a zero-shot prompting approach to VLMs such as Gemini-1.5-flash and GPT-4o-mini for road safety assessment. 
% While VLMs currently underperform compared to traditional CNNs, 
% they offer the advantage of VQA, which enables flexible prompt engineering and potential improvements in adaptability, especially valuable for resource-limited settings in LMICs.
Since VLMs enable flexible prompt engineering and result in potential adaptability improvements, which is crucial for resource-limited settings in LMICs.
% VLMs perform well on simple, interpretable attributes but struggle with spatially complex features or those requiring implicit knowledge, such as informal sidewalks. Limitations in street view coverage also hinder accurate classification, suggesting that integrating satellite imagery could provide useful context.
VLMs excel at simple, interpretable attributes but struggle with spatially complex or implicit features like informal sidewalks. Limited street view coverage also impacts accuracy, highlighting the potential benefit of integrating satellite imagery for added context.

Despite current gaps, VLMs have the potential to complement traditional methods under the iRAP framework. Future research should focus on:

\begin{itemize}
    \setlength\itemsep{0pt}
    \setlength\parskip{0pt}
    \item \textbf{Geometry reasoning:} Investigate techniques like depth estimation or segmentation to handle attributes requiring spatial measurements.
    \item \textbf{In-context learning:} Applying few-shot or Chain-of-Thought prompting for better classification.
    \item \textbf{Fine-tuning:} Adapting VLMs (e.g., with LoRA) to improve spatial and contextual understanding.
    \item \textbf{Multimodal integration:} Combining street view with satellite and GIS data to fill visual gaps.
    \item \textbf{Cross-region validation:} Testing VLMs in diverse environments for broader applicability.
    \item \textbf{Stakeholder usability:} Creating accessible tools for local authorities with minimal technical expertise.
\end{itemize}

In summary, VLMs are not yet a full substitute for trained coders but could become a scalable solution to support global road safety, particularly in under-resourced regions, through continued refinement and integration.

\section*{Acknowledgements}
We would like to thank the Thai Road Assessment Programme (ThaiRAP) and Chulalongkorn University for providing the dataset used in this work. Support for this work was provided in part by the Regan Smeed Travel Fund from the UCL Centre for Transport Studies (CTS). Natchapon Jongwiriyanurak is funded by Royal Thai Government Scholarship. Zichao Zeng, June Moh Goo and Ilya Ilyankou are funded by Ordnance Survey and  UKRI Engineering and Physical Sciences Research Council (EPSRC) [Grant numbers  EP/W522077/1, EP/X524840/1 and EP/Y528651/1]. Xinglei Wang is jointly funded by UCL Dean's Prize and China Scholarship Council [Grant number 202106270039].

{
    \small
    \bibliographystyle{ieeenat_fullname}
    \bibliography{main}

\begin{thebibliography}{49}
\providecommand{\natexlab}[1]{#1}
\providecommand{\url}[1]{\texttt{#1}}
\expandafter\ifx\csname urlstyle\endcsname\relax
  \providecommand{\doi}[1]{doi: #1}\else
  \providecommand{\doi}{doi: \begingroup \urlstyle{rm}\Url}\fi

\bibitem[Abdollahi et~al.(2020)Abdollahi, Pradhan, Shukla, Chakraborty, and Alamri]{abdollahi_deep_2020}
Abolfazl Abdollahi, Biswajeet Pradhan, Nagesh Shukla, Subrata Chakraborty, and Abdullah Alamri.
\newblock Deep {Learning} {Approaches} {Applied} to {Remote} {Sensing} {Datasets} for {Road} {Extraction}: {A} {State}-{Of}-{The}-{Art} {Review}.
\newblock \emph{Remote Sensing}, 12\penalty0 (9):\penalty0 1444, 2020.

\bibitem[Agrawal et~al.(2016)Agrawal, Lu, Antol, Mitchell, Zitnick, Batra, and Parikh]{agrawal_vqa_2016}
Aishwarya Agrawal, Jiasen Lu, Stanislaw Antol, Margaret Mitchell, C.~Lawrence Zitnick, Dhruv Batra, and Devi Parikh.
\newblock {VQA}: {Visual} {Question} {Answering}, 2016.
\newblock arXiv:1505.00468 [cs].

\bibitem[Arya et~al.(2024)Arya, Maeda, Ghosh, Toshniwal, and Sekimoto]{arya_rdd2022_2024}
Deeksha Arya, Hiroya Maeda, Sanjay~Kumar Ghosh, Durga Toshniwal, and Yoshihide Sekimoto.
\newblock {RDD2022}: {A} multi‐national image dataset for automatic road damage detection.
\newblock \emph{Geoscience Data Journal}, page gdj3.260, 2024.

\bibitem[Brkić et~al.(2020)Brkić, Miler, Ševrović, and Medak]{brkic_analytical_2020}
Ivan Brkić, Mario Miler, Marko Ševrović, and Damir Medak.
\newblock An {Analytical} {Framework} for {Accurate} {Traffic} {Flow} {Parameter} {Calculation} from {UAV} {Aerial} {Videos}.
\newblock \emph{Remote Sensing}, 12\penalty0 (22):\penalty0 3844, 2020.

\bibitem[Brkić et~al.(2022)Brkić, Miler, Ševrović, and Medak]{brkic_automatic_2022}
Ivan Brkić, Mario Miler, Marko Ševrović, and Damir Medak.
\newblock Automatic {Roadside} {Feature} {Detection} {Based} on {Lidar} {Road} {Cross} {Section} {Images}.
\newblock \emph{Sensors}, 22\penalty0 (15):\penalty0 5510, 2022.

\bibitem[Brkić et~al.(2023)Brkić, Ševrović, Medak, and Miler]{brkic_utilizing_2023}
Ivan Brkić, Marko Ševrović, Damir Medak, and Mario Miler.
\newblock Utilizing {High} {Resolution} {Satellite} {Imagery} for {Automated} {Road} {Infrastructure} {Safety} {Assessments}.
\newblock \emph{Sensors}, 23\penalty0 (9):\penalty0 4405, 2023.

\bibitem[Chen et~al.(2019)Chen, Kuhn, Prettner, and Bloom]{chen_global_2019}
Simiao Chen, Michael Kuhn, Klaus Prettner, and David~E Bloom.
\newblock The global macroeconomic burden of road injuries: estimates and projections for 166 countries.
\newblock \emph{The Lancet Planetary Health}, 3\penalty0 (9):\penalty0 e390--e398, 2019.

\bibitem[Dihan et~al.(2025)Dihan, Hassan, Parvez, Hasan, Alam, Cheema, Ali, and Parvez]{dihan_mapeval_2025}
Mahir~Labib Dihan, Md~Tanvir Hassan, Md~Tanvir Parvez, Md~Hasebul Hasan, Md~Almash Alam, Muhammad~Aamir Cheema, Mohammed~Eunus Ali, and Md~Rizwan Parvez.
\newblock {MapEval}: {A} {Map}-{Based} {Evaluation} of {Geo}-{Spatial} {Reasoning} in {Foundation} {Models}, 2025.
\newblock arXiv:2501.00316 [cs].

\bibitem[Geiger et~al.(2013)Geiger, Lenz, Stiller, and Urtasun]{geiger_vision_2013}
A Geiger, P Lenz, C Stiller, and R Urtasun.
\newblock Vision meets robotics: {The} {KITTI} dataset.
\newblock \emph{The International Journal of Robotics Research}, 32\penalty0 (11):\penalty0 1231--1237, 2013.

\bibitem[Gemini(2024)]{team_gemini_gemini_2024}
Team Gemini.
\newblock Gemini 1.5: {Unlocking} multimodal understanding across millions of tokens of context, 2024.
\newblock arXiv:2403.05530 [cs].

\bibitem[Goo et~al.(2024)Goo, Zeng, and Boehm]{goo_zero-shot_2024}
June~Moh Goo, Zichao Zeng, and Jan Boehm.
\newblock Zero-{Shot} {Detection} of {Buildings} in {Mobile} {LiDAR} using {Language} {Vision} {Model}.
\newblock \emph{The International Archives of the Photogrammetry, Remote Sensing and Spatial Information Sciences}, XLVIII-2-2024:\penalty0 107--113, 2024.

\bibitem[Goo et~al.(2025)Goo, Milidonis, Artusi, Boehm, and Ciliberto]{goo_hybrid-segmentor_2025}
June~Moh Goo, Xenios Milidonis, Alessandro Artusi, Jan Boehm, and Carlo Ciliberto.
\newblock Hybrid-{Segmentor}: {Hybrid} approach for automated fine-grained crack segmentation in civil infrastructure.
\newblock \emph{Automation in Construction}, 170:\penalty0 105960, 2025.

\bibitem[He et~al.(2015)He, Zhang, Ren, and Sun]{he_deep_2015}
Kaiming He, Xiangyu Zhang, Shaoqing Ren, and Jian Sun.
\newblock Deep {Residual} {Learning} for {Image} {Recognition}, 2015.
\newblock arXiv:1512.03385 [cs].

\bibitem[Hendrycks and Gimpel(2018)]{hendrycks_baseline_2018}
Dan Hendrycks and Kevin Gimpel.
\newblock A {Baseline} for {Detecting} {Misclassified} and {Out}-of-{Distribution} {Examples} in {Neural} {Networks}, 2018.
\newblock arXiv:1610.02136 [cs].

\bibitem[Hendrycks et~al.(2021)Hendrycks, Burns, Basart, Zou, Mazeika, Song, and Steinhardt]{hendrycks_measuring_2021}
Dan Hendrycks, Collin Burns, Steven Basart, Andy Zou, Mantas Mazeika, Dawn Song, and Jacob Steinhardt.
\newblock Measuring {Massive} {Multitask} {Language} {Understanding}, 2021.
\newblock arXiv:2009.03300 [cs].

\bibitem[Hou et~al.(2024)Hou, Quintana, Khomiakov, Yap, Ouyang, Ito, Wang, Zhao, and Biljecki]{hou_global_2024}
Yujun Hou, Matias Quintana, Maxim Khomiakov, Winston Yap, Jiani Ouyang, Koichi Ito, Zeyu Wang, Tianhong Zhao, and Filip Biljecki.
\newblock Global {Streetscapes} — {A} comprehensive dataset of 10 million street-level images across 688 cities for urban science and analytics.
\newblock \emph{ISPRS Journal of Photogrammetry and Remote Sensing}, 215:\penalty0 216--238, 2024.

\bibitem[Hudson and Manning(2019)]{hudson_gqa_2019}
Drew~A. Hudson and Christopher~D. Manning.
\newblock {GQA}: {A} {New} {Dataset} for {Real}-{World} {Visual} {Reasoning} and {Compositional} {Question} {Answering}, 2019.
\newblock arXiv:1902.09506 [cs].

\bibitem[Ibragimov et~al.(2022)Ibragimov, Lee, Lee, and Kim]{ibragimov_automated_2022}
Eldor Ibragimov, Hyun-Jong Lee, Jong-Jae Lee, and Namgyu Kim.
\newblock Automated pavement distress detection using region based convolutional neural networks.
\newblock \emph{International Journal of Pavement Engineering}, 23\penalty0 (6):\penalty0 1981--1992, 2022.

\bibitem[Ilyankou et~al.(2025)Ilyankou, Jongwiriyanurak, Cheng, and Haworth]{ilyankou_clip_2025}
Ilya Ilyankou, Natchapon Jongwiriyanurak, Tao Cheng, and James Haworth.
\newblock {CLIP} the {Landscape}: {Automated} {Tagging} of {Crowdsourced} {Landscape} {Images}, 2025.
\newblock arXiv:2506.12214 [cs].

\bibitem[Jain et~al.(2024)Jain, Thapa, Chen, Abbott, and Sarkar]{jain_semantic_2024}
Sandesh Jain, Surendrabikram Thapa, Kuan-Ting Chen, A.~Lynn Abbott, and Abhijit Sarkar.
\newblock Semantic {Understanding} of {Traffic} {Scenes} with {Large} {Vision} {Language} {Models}.
\newblock In \emph{2024 {IEEE} {Intelligent} {Vehicles} {Symposium} ({IV})}, pages 1580--1587, Jeju Island, Korea, Republic of, 2024. IEEE.

\bibitem[Jan et~al.(2018)Jan, Verma, Affum, Atabak, and Moir]{jan_convolutional_2018}
Zohaib Jan, Brijesh Verma, Joseph Affum, Sam Atabak, and Lachlan Moir.
\newblock A {Convolutional} {Neural} {Network} {Based} {Deep} {Learning} {Technique} for {Identifying} {Road} {Attributes}.
\newblock In \emph{2018 {International} {Conference} on {Image} and {Vision} {Computing} {New} {Zealand} ({IVCNZ})}, pages 1--6, Auckland, New Zealand, 2018. IEEE.

\bibitem[Jongwiriyanurak et~al.(2023)Jongwiriyanurak, Zeng, Wang, Haworth, Tanaksaranond, and Boehm]{jongwiriyanurak_framework_2023}
Natchapon Jongwiriyanurak, Zichao Zeng, Meihui Wang, James Haworth, Garavig Tanaksaranond, and Jan Boehm.
\newblock Framework for {Motorcycle} {Risk} {Assessment} {Using} {Onboard} {Panoramic} {Camera}.
\newblock In \emph{12th {International} {Conference} on {Geographic} {Information} {Science} ({GIScience} 2023)}, 2023.

\bibitem[Kacan et~al.(2020)Kacan, Orsic, Segvic, and Sevrovic]{kacan_multi-task_2020}
Marin Kacan, Marin Orsic, Sinisa Segvic, and Marko Sevrovic.
\newblock Multi-{Task} {Learning} for {iRAP} {Attribute} {Classification} and {Road} {Safety} {Assessment}.
\newblock In \emph{2020 {IEEE} 23rd {International} {Conference} on {Intelligent} {Transportation} {Systems} ({ITSC})}, pages 1--6, Rhodes, Greece, 2020. IEEE.

\bibitem[Kačan et~al.(2024)Kačan, Ševrović, and Šegvić]{kacan_dynamic_2024}
Marin Kačan, Marko Ševrović, and Siniša Šegvić.
\newblock Dynamic {Loss} {Balancing} and {Sequential} {Enhancement} for {Road}-{Safety} {Assessment} and {Traffic} {Scene} {Classification}.
\newblock \emph{IEEE Transactions on Intelligent Transportation Systems}, 25\penalty0 (11):\penalty0 15628--15640, 2024.

\bibitem[Kim et~al.(2023)Kim, Lee, and Im]{kim_multi-target_2023}
Changjae Kim, Seunghun Lee, and Sunghoon Im.
\newblock Multi-{Target} {Domain} {Adaptation} with {Class}-{Wise} {Attribute} {Transfer} in {Semantic} {Segmentation}.
\newblock In \emph{{BMVC}}, 2023.

\bibitem[Lewis et~al.(2021)Lewis, Perez, Piktus, Petroni, Karpukhin, Goyal, Küttler, Lewis, Yih, Rocktäschel, Riedel, and Kiela]{lewis_retrieval-augmented_2021}
Patrick Lewis, Ethan Perez, Aleksandra Piktus, Fabio Petroni, Vladimir Karpukhin, Naman Goyal, Heinrich Küttler, Mike Lewis, Wen-tau Yih, Tim Rocktäschel, Sebastian Riedel, and Douwe Kiela.
\newblock Retrieval-{Augmented} {Generation} for {Knowledge}-{Intensive} {NLP} {Tasks}, 2021.
\newblock arXiv:2005.11401 [cs].

\bibitem[Liang et~al.(2025)Liang, Xie, Zhao, Stouffs, and Biljecki]{liang_openfacades_2025}
Xiucheng Liang, Jinheng Xie, Tianhong Zhao, Rudi Stouffs, and Filip Biljecki.
\newblock {OpenFACADES}: {An} {Open} {Framework} for {Architectural} {Caption} and {Attribute} {Data} {Enrichment} via {Street} {View} {Imagery}, 2025.
\newblock arXiv:2504.02866 [cs].

\bibitem[Lin et~al.(2023)Lin, Tian, Duan, Zhou, Zhao, and Cao]{lin_da-rdd_2023}
Chunmian Lin, Daxin Tian, Xuting Duan, Jianshan Zhou, Dezong Zhao, and Dongpu Cao.
\newblock {DA}-{RDD}: {Toward} {Domain} {Adaptive} {Road} {Damage} {Detection} {Across} {Different} {Countries}.
\newblock \emph{IEEE Transactions on Intelligent Transportation Systems}, 24\penalty0 (3):\penalty0 3091--3103, 2023.

\bibitem[Liu et~al.(2023)Liu, Li, Wu, and Lee]{liu_visual_2023}
Haotian Liu, Chunyuan Li, Qingyang Wu, and Yong~Jae Lee.
\newblock Visual {Instruction} {Tuning}, 2023.
\newblock arXiv:2304.08485 [cs].

\bibitem[Ma et~al.(2022)Ma, Fan, Wang, Wu, Jiang, Xie, and Fan]{ma_computer_2022}
Nachuan Ma, Jiahe Fan, Wenshuo Wang, Jin Wu, Yu Jiang, Lihua Xie, and Rui Fan.
\newblock Computer vision for road imaging and pothole detection: a state-of-the-art review of systems and algorithms.
\newblock \emph{Transportation Safety and Environment}, 4\penalty0 (4):\penalty0 tdac026, 2022.

\bibitem[Marino et~al.(2019)Marino, Rastegari, Farhadi, and Mottaghi]{marino_ok-vqa_2019}
Kenneth Marino, Mohammad Rastegari, Ali Farhadi, and Roozbeh Mottaghi.
\newblock {OK}-{VQA}: {A} {Visual} {Question} {Answering} {Benchmark} {Requiring} {External} {Knowledge}, 2019.
\newblock arXiv:1906.00067 [cs].

\bibitem[OpenAI(2024{\natexlab{a}})]{openai_gpt-4_2024}
OpenAI.
\newblock {GPT}-4 {Technical} {Report}, 2024{\natexlab{a}}.
\newblock arXiv:2303.08774 [cs].

\bibitem[OpenAI(2024{\natexlab{b}})]{openai_gpt-4o_2024}
OpenAI.
\newblock {GPT}-4o {System} {Card}, 2024{\natexlab{b}}.

\bibitem[Pubudu~Sanjeewani and Verma(2019)]{pubudu_sanjeewani_learning_2019}
Thihagoda~Gamage Pubudu~Sanjeewani and Brijesh Verma.
\newblock Learning and {Analysis} of {AusRAP} {Attributes} from {Digital} {Video} {Recording} for {Road} {Safety}.
\newblock In \emph{2019 {International} {Conference} on {Image} and {Vision} {Computing} {New} {Zealand} ({IVCNZ})}, pages 1--6, Dunedin, New Zealand, 2019. IEEE.

\bibitem[Qian et~al.(2024)Qian, Chen, Zhuo, Jiao, and Jiang]{qian_nuscenes-qa_2024}
Tianwen Qian, Jingjing Chen, Linhai Zhuo, Yang Jiao, and Yu-Gang Jiang.
\newblock {NuScenes}-{QA}: {A} {Multi}-{Modal} {Visual} {Question} {Answering} {Benchmark} for {Autonomous} {Driving} {Scenario}.
\newblock \emph{Proceedings of the AAAI Conference on Artificial Intelligence}, 38\penalty0 (5):\penalty0 4542--4550, 2024.

\bibitem[Roberts(2023)]{roberts_gpt4geo_2023}
Jonathan Roberts.
\newblock {GPT4GEO}: {How} a {Language} {Model} {Sees} the {World}’s {Geography}.
\newblock In \emph{Foundation {Models} for {Decision} {Making} {Workshop} at {NeurIPS} 2023.}, 2023.

\bibitem[Sanjeewani and Verma(2021{\natexlab{a}})]{sanjeewani_optimization_2021}
Pubudu Sanjeewani and Brijesh Verma.
\newblock Optimization of {Fully} {Convolutional} {Network} for {Road} {Safety} {Attribute} {Detection}.
\newblock \emph{IEEE Access}, 9:\penalty0 120525--120536, 2021{\natexlab{a}}.

\bibitem[Sanjeewani and Verma(2021{\natexlab{b}})]{sanjeewani_single_2021}
Pubudu Sanjeewani and Brijesh Verma.
\newblock Single class detection-based deep learning approach for identification of road safety attributes.
\newblock \emph{Neural Computing and Applications}, 33\penalty0 (15):\penalty0 9691--9702, 2021{\natexlab{b}}.

\bibitem[Simonyan and Zisserman(2015)]{simonyan_very_2015}
Karen Simonyan and Andrew Zisserman.
\newblock Very {Deep} {Convolutional} {Networks} for {Large}-{Scale} {Image} {Recognition}, 2015.
\newblock arXiv:1409.1556 [cs].

\bibitem[Song et~al.(2018)Song, Workman, Hadzic, Zhang, Green, Chen, Souleyrette, and Jacobs]{song_farsa_2018}
Weilian Song, Scott Workman, Armin Hadzic, Xu Zhang, Eric Green, Mei Chen, Reginald Souleyrette, and Nathan Jacobs.
\newblock {FARSA}: {Fully} {Automated} {Roadway} {Safety} {Assessment}.
\newblock In \emph{2018 {IEEE} {Winter} {Conference} on {Applications} of {Computer} {Vision} ({WACV})}, pages 521--529, Lake Tahoe, NV, 2018. IEEE.

\bibitem[Wang et~al.(2024)Wang, Haworth, Chen, Liu, and Shi]{wang_investigating_2024}
Meihui Wang, James Haworth, Huanfa Chen, Yunzhe Liu, and Zhengxiang Shi.
\newblock Investigating the potential of crowdsourced street-level imagery in understanding the spatiotemporal dynamics of cities: {A} case study of walkability in {Inner} {London}.
\newblock \emph{Cities}, 153:\penalty0 105243, 2024.

\bibitem[Wen et~al.(2023)Wen, Yang, Fu, Wang, Cai, Li, Ma, Li, Xu, Shang, Zhu, Sun, Bai, Cai, Dou, Hu, Shi, and Qiao]{wen_road_2023}
Licheng Wen, Xuemeng Yang, Daocheng Fu, Xiaofeng Wang, Pinlong Cai, Xin Li, Tao Ma, Yingxuan Li, Linran Xu, Dengke Shang, Zheng Zhu, Shaoyan Sun, Yeqi Bai, Xinyu Cai, Min Dou, Shuanglu Hu, Botian Shi, and Yu Qiao.
\newblock On the {Road} with {GPT}-{4V}(ision): {Early} {Explorations} of {Visual}-{Language} {Model} on {Autonomous} {Driving}, 2023.
\newblock arXiv:2311.05332 [cs].

\bibitem[WHO(2023)]{who_global_2023}
WHO.
\newblock Global status report on road safety 2023.
\newblock Technical report, World Health Organization, Geneva, 2023.

\bibitem[Xing et~al.(2024)Xing, Liu, Wang, Sun, Chen, Gu, and Wang]{xing_survey_2024}
Jialu Xing, Jianping Liu, Jian Wang, Lulu Sun, Xi Chen, Xunxun Gu, and Yingfei Wang.
\newblock A survey of efficient fine-tuning methods for {Vision}-{Language} {Models} — {Prompt} and {Adapter}.
\newblock \emph{Computers \& Graphics}, 119:\penalty0 103885, 2024.

\bibitem[Xu et~al.(2025)Xu, Bai, Zhang, Li, Xia, Wong, Wang, and Zhao]{xu_drivegpt4-v2_2025}
Zhenhua Xu, Yan Bai, Yujia Zhang, Zhuoling Li, Fei Xia, Kwan-Yee~K Wong, Jianqiang Wang, and Hengshuang Zhao.
\newblock {DriveGPT4}-{V2}: {Harnessing} {Large} {Language} {Model} {Capabilities} for {Enhanced} {Closed}-{Loop} {Autonomous} {Driving}.
\newblock In \emph{{CVPR} 2025}, 2025.

\bibitem[Yin et~al.(2023)Yin, Hu, Tran, Zhang, Wang, Kruppa, Zimmermann, and Ng]{yin_multimodal_2023}
Yifang Yin, Wenmiao Hu, An Tran, Ying Zhang, Guanfeng Wang, Hannes Kruppa, Roger Zimmermann, and See-Kiong Ng.
\newblock Multimodal {Deep} {Learning} for {Robust} {Road} {Attribute} {Detection}.
\newblock \emph{ACM Transactions on Spatial Algorithms and Systems}, 9\penalty0 (4):\penalty0 1--25, 2023.

\bibitem[Yue et~al.(2024)Yue, Ni, Zhang, Zheng, Liu, Zhang, Stevens, Jiang, Ren, Sun, Wei, Yu, Yuan, Sun, Yin, Zheng, Yang, Liu, Huang, Sun, Su, and Chen]{yue_mmmu_2024}
Xiang Yue, Yuansheng Ni, Kai Zhang, Tianyu Zheng, Ruoqi Liu, Ge Zhang, Samuel Stevens, Dongfu Jiang, Weiming Ren, Yuxuan Sun, Cong Wei, Botao Yu, Ruibin Yuan, Renliang Sun, Ming Yin, Boyuan Zheng, Zhenzhu Yang, Yibo Liu, Wenhao Huang, Huan Sun, Yu Su, and Wenhu Chen.
\newblock {MMMU}: {A} {Massive} {Multi}-discipline {Multimodal} {Understanding} and {Reasoning} {Benchmark} for {Expert} {AGI}, 2024.
\newblock arXiv:2311.16502 [cs].

\bibitem[Zeng et~al.(2024)Zeng, Goo, Wang, Chi, Wang, and Boehm]{zeng_zero-shot_2024}
Zichao Zeng, June~Moh Goo, Xinglei Wang, Bin Chi, Meihui Wang, and Jan Boehm.
\newblock Zero-{Shot} {Building} {Age} {Classification} from {Facade} {Image} {Using} {GPT}-4.
\newblock \emph{The International Archives of the Photogrammetry, Remote Sensing and Spatial Information Sciences}, XLVIII-2-2024:\penalty0 457--464, 2024.

\bibitem[Zhang et~al.(2024)Zhang, Xu, and Li]{zhang_chatscene_2024}
Jiawei Zhang, Chejian Xu, and Bo Li.
\newblock {ChatScene}: {Knowledge}-{Enabled} {Safety}-{Critical} {Scenario} {Generation} for {Autonomous} {Vehicles}.
\newblock In \emph{2024 {IEEE}/{CVF} {Conference} on {Computer} {Vision} and {Pattern} {Recognition} ({CVPR})}, pages 15459--15469, Seattle, WA, USA, 2024. IEEE.

\end{thebibliography}
}

% \clearpage

\section*{Supplemental Material}
% \section*{Attribute-level Performance Comparison}

% \renewcommand{\arraystretch}{1.2}
\begin{table*}
\centering
% \caption{Performance comparison across all iRAP-defined attribute groups using four models: VGG, Res(Net), GPT(-4o-mini), and Gem(ini-1.5-flash). The following abbreviations are used: (OF) = Observed Flow, (F) = Facilities, (DS) = Driver-Side, (PS) = Passenger-Side, (PC) = Pedestrian Crossing, (POF) = Pedestrian Observed Flow, (RS) = Roadside Severity, (SZ) = School Zone. Metrics include Accuracy, Precision, Recall, and F1-score. For each attribute group, the last row (“All”) shows the macro-average across attributes within the group. The best-performing model in each metric is {bolded}, while the best-performing vision–language model (VLM) is \underline{underlined}.}
\caption{Performance comparison across all iRAP-defined attributes using four models: VGG, Res(Net), GPT(-4o-mini), and Gem(ini-1.5-flash). Attribute group abbreviations: OF = Observed Flow, F = Facilities, DS = Driver-Side, PS = Passenger-Side, PC = Pedestrian Crossing, POF = Pedestrian Observed Flow, RS = Roadside Severity, SZ = School Zone. Metrics include Accuracy, Precision, Recall, and F1-score. For each attribute group, the last row (“All”) reports the macro-average across all attributes. \textbf{Bold} indicates the best-performing model.}
\label{tab:attr_performance_full}
\resizebox{\textwidth}{!}{%
\begin{tabular}{llcccccccccccccccc}
\hline
\multirow{3}{*}{Group Attribute} & \multirow{3}{*}{Attribute} 
& \multicolumn{4}{c}{Accuracy} 
& \multicolumn{4}{c}{Precision} 
& \multicolumn{4}{c}{Recall} 
& \multicolumn{4}{c}{F1} \\
\cline{3-6} \cline{7-10} \cline{11-14} \cline{15-18}
& & \multicolumn{2}{c}{CNN} & \multicolumn{2}{c}{VLM} 
& \multicolumn{2}{c}{CNN} & \multicolumn{2}{c}{VLM} 
& \multicolumn{2}{c}{CNN} & \multicolumn{2}{c}{VLM} 
 & \multicolumn{2}{c}{CNN} & \multicolumn{2}{c}{VLM} \\
\cline{3-4} \cline{5-6} \cline{7-8} \cline{9-10} 
\cline{11-12} \cline{13-14} \cline{15-16} \cline{17-18}
& & VGG & Res & GPT & Gem 
  & VGG & Res & GPT & Gem 
  & VGG & Res & GPT & Gem 
  & VGG & Res & GPT & Gem \\
\hline
\multirow{6}{*}{\rotatebox{90}{Observed    Flow}}       & Motorcycle (OF)                                & 0.96       & 0.96          & 0.80 & 0.93          & 0.48 & 0.67          & 0.20       & 0.58          & 0.48 & 0.97          & 0.27       & 0.42          & 0.48 & 0.73          & 0.20       & 0.46          \\
                                        & Bicycle (OF)                                   &            &               & 1.00 & 1.00          &      &               & 1.00       & 1.00          &      &               & 1.00       & 1.00          &      &               & 1.00       & 1.00          \\
                                        & (POF) across the road                          &            &               & 1.00 & 1.00          &      &               & 0.50       & 1.00          &      &               & 0.50       & 1.00          &      &               & 0.50       & 1.00          \\
                                        & (POF) along the road (DS)                      & 1.00       & 1.00          & 0.99 & 0.99          & 1.00 & 1.00          & 0.33       & 0.33          & 1.00 & 1.00          & 0.33       & 0.33          & 1.00 & 1.00          & 0.33       & 0.33          \\
                                        & (POF) along the road (PS)                      & 1.00       & 1.00          & 0.99 & 0.99          & 1.00 & 1.00          & 0.33       & 0.50          & 1.00 & 1.00          & 0.33       & 0.50          & 1.00 & 1.00          & 0.33       & 0.50          \\
                                        & {All Observed    Flow}                                         & \textbf{ 0.99} & \textbf{ 0.99}    & 0.95 & {0.98} & 0.83 & \textbf{ 0.89}    & 0.47       & {0.68} & 0.83 & \textbf{ 0.99}    & 0.49       & {0.65} & 0.83 & \textbf{ 0.91}    & 0.47       & {0.66} \\ \hline

\multirow{6}{*}{\rotatebox{90}{Speed Limits}}           & Speed limit                                    & 0.99       & 0.98          & 0.44 & 0.74          & 0.65 & 0.98          & 0.34       & 0.50          & 0.67 & 0.97          & 0.33       & 0.48          & 0.66 & 0.98          & 0.30       & 0.48          \\
                                        & Motorcycle speed limit                         & 1.00       & 1.00          & 0.54 & 0.96          & 0.48 & 1.00          & 0.55       & 0.86          & 0.50 & 1.00          & 0.76       & 0.73          & 0.49 & 1.00          & 0.44       & 0.78          \\
                                        & Truck speed limit                              & 1.00       & 1.00          & 0.62 & 0.94          & 0.48 & 1.00          & 0.38       & 0.57          & 0.50 & 1.00          & 0.52       & 0.48          & 0.49 & 1.00          & 0.34       & 0.52          \\
                                        & Differential speed limits                      & 0.99       & 0.98          & 0.21 & 0.74          & 1.00 & 0.97          & 0.60       & 0.39          & 1.00 & 0.96          & 0.51       & 0.46          & 1.00 & 0.97          & 0.19       & 0.43          \\
                                        & Speed management                               & 0.98       & 0.96          & 0.80 & 0.80          & 1.00 & 0.97          & 0.40       & 0.40          & 1.00 & 0.96          & 0.50       & 0.50          & 1.00 & 0.97          & 0.45       & 0.45          \\
                                        & {All Speed Limits}                                            & \textbf{ 0.99} & 0.98          & 0.52 & {0.84} & 0.72 & \textbf{ 0.98}    & 0.46       & {0.55} & 0.73 & \textbf{ 0.98}    & 0.52       & {0.53} & 0.73 & \textbf{ 0.98}    & 0.34       & {0.53} \\ \hline
\multirow{17}{*}{\rotatebox{90}{Mid-block}} & Number of lanes                                & 0.95       & 0.95          & 0.14 & 0.55          & 0.66 & 0.82          & 0.29       & 0.30          & 0.44 & 0.82          & 0.32       & 0.54          & 0.49 & 0.82          & 0.09       & 0.26          \\
                                        & Lane width                                     &            &               & 1.00 & 1.00          &      &               & 1.00       & 1.00          &      &               & 1.00       & 1.00          &      &               & 1.00       & 1.00          \\
                                        & Curvature                                      & 0.99       & 0.99          & 0.98 & 0.98          & 0.50 & 1.00          & 0.49       & 0.62          & 0.48 & 1.00          & 0.50       & 0.62          & 0.49 & 1.00          & 0.49       & 0.62          \\
                                        & Quality of curve                               & 0.99       & 0.99          & 0.98 & 0.98          & 0.50 & 1.00          & 0.49       & 0.62          & 0.48 & 1.00          & 0.50       & 0.62          & 0.49 & 1.00          & 0.49       & 0.62          \\
                                        & Upgrade cost                                   & 0.93       & 0.96          & 0.49 & 0.61          & 0.92 & 0.98          & 0.57       & 0.39          & 0.83 & 0.97          & 0.34       & 0.45          & 0.86 & 0.97          & 0.25       & 0.40          \\
                                        & Median type                                    & 0.93       & 0.95          & 0.19 & 0.72          & 0.67 & 0.97          & 0.17       & 0.21          & 0.70 & 0.84          & 0.22       & 0.25          & 0.68 & 0.89          & 0.08       & 0.23          \\
                                        & Skid resistance / grip                         & 0.98       & 0.99          & 0.80 & 0.80          & 0.98 & 0.98          & 0.40       & 0.73          & 0.95 & 0.95          & 0.50       & 0.51          & 0.96 & 0.96          & 0.44       & 0.47          \\
                                        & Road condition                                 & 0.00       & 0.00          & 0.99 & 0.98          &      &               & 0.50       & 0.50          &      &               & 0.50       & 0.49          &      &               & 0.50       & 0.49          \\
                                        & Vehicle parking                                & 0.99       & 1.00          & 0.79 & 0.97          & 0.48 & 0.48          & 0.33       & 0.50          & 0.50 & 0.50          & 0.27       & 0.49          & 0.49 & 0.49          & 0.30       & 0.49          \\
                                        & Grade                                          &            &               & 1.00 & 1.00          &      &               & 1.00       & 1.00          &      &               & 1.00       & 1.00          &      &               & 1.00       & 1.00          \\
                                        & Roadworks                                      &            &               & 1.00 & 0.99          &      &               & 0.50       & 0.50          &      &               & 0.50       & 0.50          &      &               & 0.50       & 0.50          \\
                                        & Sight distance                                 &            &               & 1.00 & 1.00          &      &               & 1.00       & 1.00          &      &               & 1.00       & 1.00          &      &               & 1.00       & 1.00          \\
                                        & Delineation                                    & 1.00       & 1.00          & 1.00 & 1.00          & 0.50 & 1.00          & 0.50       & 0.50          & 0.48 & 1.00          & 0.50       & 0.50          & 0.49 & 1.00          & 0.50       & 0.50          \\
                                        & Street lighting                                & 0.96       & 0.97          & 0.89 & 0.94          & 1.00 & 1.00          & 0.62       & 0.75          & 1.00 & 1.00          & 0.68       & 0.72          & 1.00 & 1.00          & 0.64       & 0.73          \\
                                        & Service road                                   & 0.97       & 0.98          & 0.12 & 0.10          & 0.47 & 0.98          & 0.55       & 0.05          & 0.50 & 0.75          & 0.51       & 0.50          & 0.48 & 0.83          & 0.11       & 0.09          \\
                                        & Centreline rumble strips                       &            &               & 1.00 & 1.00          &      &               & 1.00       & 1.00          &      &               & 1.00       & 1.00          &      &               & 1.00       & 1.00          \\ 
                                        & {All Mid-block}                                            & 0.97       & \textbf{ 0.98}    & 0.77 & {0.85} & 0.67 & \textbf{ 0.92}    & 0.59       & {0.60} & 0.64 & \textbf{ 0.88} & 0.58       & {0.64}    & 0.64 & \textbf{ 0.90} & 0.52       & {0.59}    \\ \hline
\multirow{8}{*}{\rotatebox{90}{Roadside}}    & (RS) - (DS) distance                           & 0.97       & 0.96          & 0.13 & 0.00          & 0.91 & 0.94          & 0.25       & 0.00          & 0.91 & 0.98          & 0.17       & 0.00          & 0.91 & 0.96          & 0.10       & 0.00          \\
                                        & (RS) - (DS) object                             & 0.93       & 0.94          & 0.25 & 0.82          & 0.46 & 0.46          & 0.18       & 0.27          & 0.48 & 0.48          & 0.10       & 0.29          & 0.47 & 0.47          & 0.10       & 0.28          \\
                                        & (RS) - (PS) distance                           & 0.86       & 0.88          & 0.36 & 0.04          & 0.60 & 0.66          & 0.39       & 0.03          & 0.58 & 0.52          & 0.38       & 0.26          & 0.57 & 0.55          & 0.22       & 0.04          \\
                                        & (RS) - (PS) object                             & 0.71       & 0.76          & 0.24 & 0.26          & 0.82 & 0.89          & 0.29       & 0.06          & 0.80 & 0.87          & 0.26       & 0.16          & 0.80 & 0.88          & 0.16       & 0.09          \\
                                        & Shoulder rumble strips                         & 0.99       & 0.98          & 0.99 & 0.96          & 0.48 & 0.48          & 0.50       & 0.50          & 0.50 & 0.50          & 0.50       & 0.48          & 0.49 & 0.49          & 0.50       & 0.49          \\
                                        & Paved shoulder - (DS)                          & 0.98       & 0.96          & 0.11 & 0.02          & 0.91 & 0.88          & 0.22       & 0.25          & 0.91 & 0.84          & 0.03       & 0.01          & 0.91 & 0.86          & 0.06       & 0.01          \\
                                        & Paved shoulder - (PS)                          & 0.87       & 0.89          & 0.19 & 0.41          & 0.86 & 0.98          & 0.09       & 0.31          & 0.87 & 0.94          & 0.11       & 0.27          & 0.86 & 0.96          & 0.10       & 0.25          \\
                                        & {All Roadside}                                           & 0.90       & \textbf{ 0.91} & 0.33 & {0.36}    & 0.72 & \textbf{ 0.76} & {0.27} & 0.20          & 0.72 & \textbf{ 0.73} & {0.22} & 0.21          & 0.72 & \textbf{ 0.74} & {0.18} & 0.17          \\ \hline
\multirow{14}{*}{\rotatebox{90}{Intersections}}         & Land use - (DS)                                & 1.00       & 1.00          & 0.38 & 0.99          & 1.00 & 1.00          & 0.33       & 0.33          & 1.00 & 1.00          & 0.46       & 0.33          & 1.00 & 1.00          & 0.19       & 0.33          \\
                                        & Land use - (PS)                                & 0.86       & 0.88          & 0.70 & 0.26          & 0.55 & 0.55          & 0.26       & 0.21          & 0.53 & 0.54          & 0.31       & 0.17          & 0.54 & 0.54          & 0.27       & 0.11          \\
                                        & Area type                                      & 0.96       & 0.98          & 0.90 & 0.82          & 1.00 & 1.00          & 0.68       & 0.66          & 1.00 & 1.00          & 0.58       & 0.86          & 1.00 & 1.00          & 0.61       & 0.68          \\
                                        & (PC) (F) - inspected road                      & 0.94       & 0.96          & 0.94 & 0.95          & 0.48 & 0.48          & 0.31       & 0.57          & 0.48 & 0.47          & 0.33       & 0.39          & 0.48 & 0.48          & 0.32       & 0.42          \\
                                        & (PC) (F) - intersecting road                   & 0.99       & 0.99          & 0.99 & 0.99          & 1.00 & 1.00          & 0.33       & 0.33          & 1.00 & 1.00          & 0.33       & 0.33          & 1.00 & 1.00          & 0.33       & 0.33          \\
                                        & (PC) quality                                   & 0.95       & 0.95          & 0.93 & 0.93          & 0.48 & 0.48          & 0.47       & 0.47          & 0.48 & 0.47          & 0.50       & 0.50          & 0.48 & 0.48          & 0.48       & 0.48          \\
                                        & Pedestrian fencing                             & 1.00       & 1.00          & 1.00 & 1.00          & 1.00 & 1.00          & 0.50       & 0.50          & 1.00 & 1.00          & 0.50       & 0.50          & 1.00 & 1.00          & 0.50       & 0.50          \\
                                        & Sidewalk - (DS)                                &            &               & 0.89 & 0.98          &      &               & 0.50       & 0.33          &      &               & 0.45       & 0.33          &      &               & 0.47       & 0.33          \\
                                        & Sidewalk - (PS)                                & 0.92       & 0.93          & 0.77 & 0.83          & 0.34 & 0.72          & 0.18       & 0.31          & 0.32 & 0.74          & 0.21       & 0.20          & 0.33 & 0.73          & 0.20       & 0.20          \\
                                        & (F) for bicycles                               &            &               & 1.00 & 1.00          &      &               & 1.00       & 1.00          &      &               & 1.00       & 1.00          &      &               & 1.00       & 1.00          \\
                                        & (F) for motorised two wheelers                 &            &               & 1.00 & 1.00          &      &               & 0.50       & 1.00          &      &               & 0.50       & 1.00          &      &               & 0.50       & 1.00          \\
                                        & (SZ) crossing supervisor                       & 0.96       & 0.98          & 0.96 & 0.96          & 0.73 & 0.98          & 0.48       & 0.32          & 0.73 & 0.75          & 0.50       & 0.33          & 0.73 & 0.83          & 0.49       & 0.33          \\
                                        & (SZ) warning                                   & 0.96       & 0.98          & 0.96 & 0.96          & 0.98 & 0.98          & 0.48       & 0.32          & 0.75 & 0.75          & 0.50       & 0.33          & 0.83 & 0.83          & 0.49       & 0.33          \\
                                        & {All Intersections}                                            & 0.95       & \textbf{ 0.97} & 0.88 & {0.90}    & 0.76 & \textbf{ 0.82} & 0.46       &  {0.49}    & 0.73 & \textbf{ 0.77} &  {0.48} &  {0.48}    & 0.74 & \textbf{ 0.79} & 0.45       &  {0.46} \\ \hline
\end{tabular}}
\end{table*}

This supplementary section expands on the attribute-level analysis introduced in Section 5.2.3 of the main paper. Table~\ref{tab:attr_performance_full} provides detailed performance results for all 52 iRAP-defined attributes, comparing four models, two CNN-based baselines (VGG and ResNet) and two vision–language models (GPT-4o-mini and Gemini-1.5-flash). 

Metrics reported include Accuracy, Precision, Recall, and F1-score. The best-performing model for each metric is \textbf{bolded}. This breakdown complements the summary metrics in the main paper and highlights performance variation across diverse attribute types.

As discussed, vision–language models are capable of zero-shot prediction across all attributes, including those with limited or no training samples in the CNN setting.

% \section*{1. Extended Performance Comparison}

% \noindent
% \begin{minipage}[t]{0.48\textwidth}
% This supplementary section expands on the attribute-level analysis introduced in Section~4.3 of the main paper. Table~3 provides detailed performance results for all 52 iRAP-defined attributes, comparing four models: two CNN-based baselines (VGG and ResNet) and two vision--language models (GPT-4o-mini and Gemini-1.5-flash).
% \end{minipage}%
% \hfill
% \begin{minipage}[t]{0.48\textwidth}
% Metrics reported include Accuracy, Precision, Recall, and F1-score. The best-performing model for each metric is \textbf{bolded}. This breakdown complements the summary metrics in the main paper and highlights performance variation across diverse attribute types.

% As discussed, vision--language models are capable of zero-shot prediction across all attributes, including those with limited or no training samples in the CNN setting.
% \end{minipage}

% \vspace{1em}

\end{document}